\documentclass[journal, twoside]{IEEEtran}  
\IEEEoverridecommandlockouts                              

\let\labelindent\relax 
\usepackage[inline]{enumitem}
\usepackage{mathtools} 
\usepackage{amssymb}  
\usepackage{graphicx}
\usepackage{siunitx} 
\usepackage{balance}
\usepackage{url}
\usepackage{bm}
\usepackage[normalem]{ulem} 
\usepackage[makeroom]{cancel} 
\usepackage[dvipsnames]{xcolor}
\usepackage[font=small,skip=2pt,position=auto,belowskip=-0.5em]{caption}
\usepackage{subcaption}
\usepackage{rotating}
\usepackage[linesnumbered, ruled, vlined]{algorithm2e}
\usepackage[colorlinks,bookmarksopen,bookmarksnumbered,citecolor=black,urlcolor=black]{hyperref}
\usepackage{booktabs} 
\usepackage{pifont} 
\hypersetup
{
	pdfauthor = {Josep Martí­-Saumell},
	pdfsubject = {Submitted to Transactions on Robotics},
  pdfkeywords = {aerial manipulation, model predictive control, optimal control, differential dynamic programming},
	pdftoolbar = true,
	colorlinks = true,
	linkcolor = black,
	citecolor = black,
	urlcolor = black,
}
\usepackage{soul}
\usepackage{glossaries}
\newacronym[firstplural=unmanned aerial manipulators (UAMs)]{am}{UAM}{unmanned aerial manipulator}
\newacronym{nmpc}{nMPC}{non-linear model predictive control}
\newacronym[firstplural=degrees of freedom (DOFs)]{dof}{DoF}{degree of freedom}
\newacronym{aba}{ABA}{Articulated Body Algorithm}
\newacronym{fd}{FD}{forward dynamics}
\newacronym{pmp}{PMP}{Pontryagin's Minimum Principle}
\newacronym{hjb}{HJB}{Hamilton-Jacobi-Bellman}
\newacronym{nlp}{NLP}{non-linear programming}
\newacronym{ik}{IK}{inverse kinematics}
\newacronym{id}{ID}{inverse dynamics}
\newacronym{ddp}{DDP}{Differential Dynamic Programming}
\newacronym{fddp}{FDDP}{Feasibility-prone Differential Dynamic Programming}
\newacronym{esc}{ESC}{electronic speed controller}
\newacronym{to}{TO}{trajectory optimization}
\newacronym{kkt}{KKT}{Karush-Kuhn-Tucker}
\newacronym{ilqr}{iLQR}{iterative linear quadratic regulator}
\newacronym{slq}{SLQ}{sequential linear quadratic}
\newacronym{tpbvp}{TPBVP}{two point boundary value problem}
\newacronym{ocp}{OCP}{optimal control problem}
\newacronym{qp}{QP}{quadratic programming}
\newacronym{rk}{RK}{Runge-Kutta}
\newacronym{wmpc}{W-MPC}{weighted MPC}
\newacronym{rmpc}{R-MPC}{rail MPC}
\newacronym{cmpc}{C-MPC}{carrot MPC}
\newacronym{ode}{ODE}{ordinary differential equation}
\newacronym{uaw}{UAW}{underactuation-aware weighting}
\newlength{\tempdima}
\newcommand{\rowname}[1]{\rotatebox{0}{\makebox[\tempdima][c]{(#1)}}}
\newcommand{\cmark}{\ding{51}}%

\title{\LARGE \bf Full-Body Torque-Level Non-linear Model Predictive Control\\ for Aerial Manipulation}

\author{Josep Mart\'i-Saumell \quad Joan Sol\`a  \quad  Angel Santamaria-Navarro \quad  Juan Andrade-Cetto
\thanks{J. Mart\'i-Saumell, J. Sol\`a  and J. Andrade-Cetto are with the Institut de Rob\`otica i Inform\`atica Industrial, CSIC-UPC, Llorens Artigas 4-6, Barcelona 08028 (e-mail: \{jmarti, jsola, cetto\}@iri.upc.edu).}%
\thanks{A. Santamaria-Navarro is with the NASA-Jet Propulsion Laboratory, California Institute of Technology, Pasadena, CA 91109 USA (e-mail: angel.santamaria.navarro@jpl.nasa.gov).}
\thanks{This work was partially supported by the EU H2020 project GAUSS (H2020-Galileo-2017-1-776293), project EB-SLAM (DPI2017-89564-P), by the Spanish State Research Agency through the Mar\'ia de Maeztu Seal of Excellence to IRI (MDM-2016-0656). Part of this research was carried out at the Jet Propulsion Laboratory, California Institute of Technology, under a contract with the National Aeronautics and Space Administration. U.S. Government sponsorship acknowledged.}
}

\graphicspath{ {./figures/} }

\begin{document}

\maketitle
\thispagestyle{plain}
\pagestyle{plain}

\newcommand{\phtopic}[1]{{\bf#1\\}} 
\newcommand{\ie}{\emph{i.e.}}
\newcommand{\eg}{\emph{e.g.}}
\newcommand{\tdot}[1]{\dot{\tilde{#1}}}
\newcommand{\com}[1]{{\color{red}#1}}
\newcommand{\comb}[1]{{\color{blue}#1}}
\newcommand{\como}[1]{{\color{orange}#1}}
\newcommand{\stb}[1]{{\color{blue}\st{#1}}}
\newcommand{\figref}[1]{Fig.~\ref{#1}}
\newcommand{\secref}[1]{Section~\ref{#1}}
\newcommand{\algref}[1]{Algorithm~\ref{#1}}
\newcommand{\tabref}[1]{Table~\ref{#1}}
\newcommand{\dt}{\delta t}
\newcommand{\Dt}{\Delta t}

\begin{abstract}
Non-linear model predictive control (nMPC) is a powerful approach to control complex robots (such as  humanoids, quadrupeds or unmanned aerial manipulators (UAMs)) as it brings important advantages over other existing techniques.
The full-body dynamics, along with the prediction capability of the optimal control problem (OCP) solved at the core of the controller, allows to actuate the robot in line with its dynamics.
This fact enhances the robot capabilities and allows, \eg, to perform intricate maneuvers at high dynamics while optimizing the amount of energy used.
Despite the many similarities between humanoids or quadrupeds and UAMs, full-body torque-level nMPC has rarely been applied to UAMs.

This paper provides a thorough description of how to use such techniques in the field of aerial manipulation.
We give a detailed explanation of the different parts involved in the OCP, from the UAM dynamical model to the residuals in the cost function.
We develop and compare three different nMPC controllers: Weighted MPC, Rail MPC and Carrot MPC, which differ on the structure of their OCPs and on how these are updated at every time step.
To validate the proposed framework, we present a wide variety of simulated case studies. 
First, we evaluate the trajectory generation problem, \ie, optimal control problems solved offline, involving different kinds of motions (\eg, aggressive maneuvers or contact locomotion) for different types UAMs. 
Then, we assess the performance of the three nMPC controllers, \ie, closed-loop controllers solved online, through a variety of realistic  simulations.
For the benefit of the community, we have made available the source code related to this work.

\end{abstract}

\begin{IEEEkeywords}
	unmanned aerial manipulation, optimal control, non-linear model predictive control, differential dynamic programming, locomotion
\end{IEEEkeywords}


\section{Introduction} \label{sec:introduction}
\begin{figure}[t]
  \centering
  \href{https://youtu.be/I-32ni4S2wg}{\includegraphics[width=\linewidth]{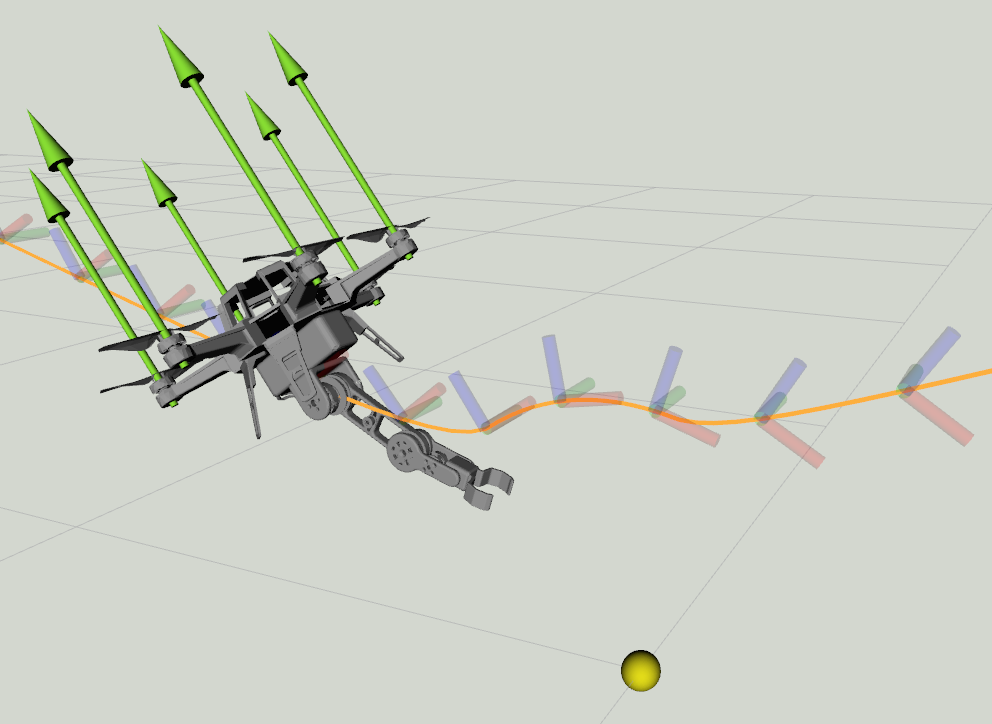}}
  \caption{Hexacopter with a 3DoF arm about to catch the ball in a non-stop flight maneuver, the \emph{Eagle's Catch} maneuver, driven by a full-body, torque-level, non-linear model predictive controller. See the video of the paper at: https://youtu.be/I-32ni4S2wg. 
  }
  \label{fig:cover}
\end{figure}

An \gls*{am} is in many ways very similar to a humanoid.
Both robots have arms attached to bodies that can move around.
They can push, grasp, manipulate, catch, move, turn, throw, pull and transport objects; they may lean, rest, jump, or draw on surfaces; or wave, signal or point at things or people.
They always have to fight against gravity, as their platforms are naturally unstable and constantly tend to lose balance and fall.
They want to be compliant with unexpected or inaccurately planned contacts, so as to keep the robot integrity.
They are redundant, meaning that they can accomplish their tasks in different manners.
They are underactuated, implying that they often have to plan complicated trajectories, with tricky dynamic maneuvers, to reach their goals (\figref{fig:cover}).

For all these reasons, these robots are similarly difficult to control, and the kinds of challenges they present are almost parallel.
Yet the control approaches observed in the respective literature bodies are significantly different.
One could think this is precisely because of the differences between them: flying versus walking. 
But we believe that these differences do not justify the diverging approaches.
After all, we could add  propellers to a humanoid, or make a multi-arm \gls{am} walk or jump.
At least at the conceptual level, what would then be the differences between these platforms?
They would clearly vanish, suggesting a convergence of the control methods.

There are several aspects of motion control that are relevant to \glspl{am}. 
Among others, we can highlight the following dichotomies, where the second options seem fairly desirable: separate base and arm controls vs. full-body control; kinematics vs. dynamics (\ie, position or velocity control vs. torque control); and non-predictive vs. predictive control. 
The humanoids and the legged robots communities have undergone all these steps and converged to strategies of full-body optimal predictive control at the torque level.
These formulations are at the same time generic, intuitive and powerful, have demonstrated their fitness to board many of the control challenges for such complex robots~\cite{carpentier_leggedrobots_2021}, especially when high dynamics are involved.
In contrast, we have not seen these methods used in \glspl{am}. 

In view of this, in this paper we walk through the process of adapting the full-body, torque-level, optimal control methods~\cite{carpentier_leggedrobots_2021,betts_practicalmethods_optimalcontrol_book}, used in the communities of humanoids and legged robots to the field of \glspl*{am}.
The use of full-body \gls*{nmpc} simplifies the control architecture since it unifies several functional blocks.
There are other advantages associated with using full-body \gls{nmpc}.
First, its genericity enables its usage by a  wide variety of \glspl*{am} (\ie, diverse multicopter platforms and manipulators) in a wide variety of tasks.
The numerical approach to the problem allows to change the \gls*{am} with a small overhead on the modeling side, which is mostly automated.
The control objectives as well as the constraints imposed by the system or the problem can be specified intuitively through cost functions and constraints in the \gls*{ocp} residing at the core of the \gls*{nmpc}.
Besides, the prediction capability of this \gls*{ocp} allows to create and execute fast and aggressive motions that are in line with the dynamics of the robot.
These dynamics can be properly exploited thanks to the force/torque control, thereby enabling the execution of tasks that would not be feasible statically.
Finally, hybrid problems like walking  or the pick-and-place operation (involving the change of the dynamic parameters) are easily modeled and handled with the optimal control approach.

Today, the \gls*{ocp} required in the \gls*{nmpc} techniques can be solved online.
This is thanks to recent advances in fast non-linear solvers and efficient methods to numerically evaluate the dynamics of complex mechanical systems (such as legged robots, humanoids and \glspl{am}).
Of course, adapting the stated methods and implementing the related software does not suffice to enhance the capabilities of the current \glspl*{am}.
In particular, progress on the hardware side is also required, \ie, we need torque-controlled, rigid, and lightweight arms to mount on the flying platforms.
Though we do not cover these aspects in the present paper, we do show through realistic simulations that the required electro-mechanic specifications can be achieved with currently available inexpensive hardware (especially motors, power electronics and low-weight CPUs) and mechanics (composite or 3D-printed arms).

\subsection{Contribution \& Paper outline}
\label{subsec:contribution}

The contributions of this work include:
\begin{enumerate}
\item A thorough description to the approach, including the theory behind fast \gls{ocp} solvers, the specification of the cost functions and the modeling of the dynamics.
  \item Carrot-\gls{nmpc}, a new strategy to specify the \gls*{ocp} in the \gls*{nmpc} that fully respects the optimality criteria used to define the mission. This is compared against previous approaches  adapted from other fields of robotics.
  \item A thorough validation through realistic simulation over a variety of platforms, arms and tasks, eventually involving high dynamics, providing evidence of computing times and hardware requirements that should allow real implementations in the immediate future.
  \item All the software tools used to conduct the presented experiments in \secref{sec:validation} are made available for the benefit of the community.
\end{enumerate}

The rest of the paper is structured as follows.
In \secref{sec:soa}, we give an overview of the state-of-the-art techniques related to the control of \glspl*{am}.
In \secref{sec:optimal_control}, we present the principal methods to solve \glspl*{ocp}, giving special attention to the fast solvers used in this work.
The following two sections are devoted to explaining the formulation of the several parts involved in the \gls*{ocp}.
In \secref{sec:modeling} we explain how we build the dynamical model of the \gls*{am}, including the case where it is in contact with the environment.
Then in \secref{sec:cost_functions} we show how we build the cost function by describing its structure as well as its different residual functions.
In \secref{sec:mpc} we present different strategies to move from the \gls*{ocp} used to generate trajectories offline, to online \glspl*{ocp} used to control the robots (\gls*{nmpc}).
The described methods are finally tested in different experiments in \secref{sec:validation}. 
First, we present a set of \gls{am} missions involving complex situations. 
These are solved offline to generate optimal state and control trajectories.
Second, we compare the performance of different \gls*{nmpc} techniques for several platforms and tasks in a realistic real-time simulated environment.
The paper closes in \secref{sec:conclusions} with our conclusions.


\section{State of the art} \label{sec:soa}
\subsection{Motion control}
Motion controllers for \glspl{am} can be divided into two categories~\cite{ruggiero_am_literature_review_2018,khamseh_am_survey_2018}.
On the one hand, \emph{decentralized controllers} control separately the platform and the robotic arm~\cite{ruggiero_multilayer_2015}.
Hence, the respective controllers treat the interaction force as a disturbance.
These methods are suited when the dynamics of the arm are insignificant, which makes the interaction force to be small.
This can be achieved either by considering lightweight arms or by executing slow movements (\eg, assuming zero roll and pitch angles of the base platform).
A typical scheme for a decentralized approach is to consider a motion planner at the kinematic level, which normally resolves the kinematic redundancy of the \gls*{am}~\cite{rossi_trajgeneration_2017,lipiello_visualservoing_2016} (\eg, employing a hierarchical closed-loop inverse kinematic algorithm).
Since it is common to have \glspl*{am} with arms controlled by position, these algorithms output conservative position set-points for the lower controllers of the platform and the arm, not only to minimize the interaction force but also to ensure that each controller reaches the desired positions within the required time.

On the other hand, \emph{centralized controllers} do full-body control of the \gls*{am}.
These are well-suited when the dynamics of the arm are significant.
Among them, it is a common choice to apply feedback linearization.
With fully-actuated platforms (like the one presented in~\cite{rajappa_tilthex_2015}), we can design a control law that fully linearizes the dynamics of the system~\cite{tognon_control-aware_2018}.
With under-actuated platforms, we can still linearize a portion of the dynamics by carefully choosing the output variable~\cite{kim_stabilizing_2018}.
In any case, with the linearized (canceled) dynamics, the problem reduces to designing proper reference trajectories for the selected outputs. 
By assigning the proper dynamics to these outputs, one can create the output trajectories needed by the linearizing controller to drive the outputs to the desired value.

Alternatively, for \emph{differentially flat}\footnote{Differential flatness is a property of a system that allows us to express the states and inputs as a function of several outputs and their higher order derivatives. These are the so-called \emph{flat outputs}} 
systems, one could generate a sufficiently smooth trajectory  in the \emph{flat output} space.
This ensures that the system is able to follow the planned trajectory.
Differential flatness implies feedback linearizability, which eases the application of a linearizing controller (see~\cite{murray_differentialflatness_1995,fliess_differentialflatness_1999} for further details on this equivalence).
Proving that several outputs constitute a set of flat outputs involves manipulating the equations of the dynamics of the system.
In aerial robotics, differential flatness was initially used for simple multicopter-based systems~\cite{mellinger_minimum_2011,sreenath_cablesuspended_2013}.
When the complexity of the system increases, which is the case of \glspl*{am}, finding diferentially flat outputs becomes a difficult task.
In \cite{yuksel_differential_2016,tognon_dynamic_2017} it is shown how a \gls*{am} can be differentially flat.
However, this only holds if the first joint of the arm coincides with the center of mass of the platform.
Moreover, the arm is restricted to be planar, \ie, all the axes of the joints should be parallel, which severely limits its practical interest.


\subsection{Environment interaction}
Any \gls*{am} control technique should enable a fundamental capability: the interaction with the environment.
In~\cite{ryll_6dinteraction_2019}, they consider a fully-actuated hexacopter with a rigidly attached end-effector to solve several problems involving physical interaction.
Ref.~\cite{hamaza_adaptivecompliance_2018} proposes a \gls*{am} design with a 1-\gls*{dof} arm to apply a force to a specific point.
In~\cite{tognon_trulyredundant_2019}, a fully-actuated platform with a 2-\gls*{dof} arm is used to perform a ``push \& slide" maneuver in a cylindrical surface.
The interaction considered in these approaches is limited to the application of forces on surfaces.
However, the idea of leveraging the contact to create motion (\ie, the contact-locomotion for \gls*{am}) has been rarely explored and, to the best of the authors' knowledge, there exists only one reference exploring this path,~\cite{delamare_toward_2018}, with a simplified 2D model of an \gls*{am} (\eg, with a 1-\gls*{dof} arm) hanging from a bar that is able to jump to another bar by exploiting the gripping contact.

Another common situation where \glspl*{am} interact with the environment involves load transportation pick and place operations.
When catching or releasing the load, one should account for the changes in the dynamic parameters involved in these operations (addition of mass and change in the inertia matrix of one of the arm links).

\subsection{Optimal control in aerial manipulation}
There have been several works applying optimal control to aerial manipulation problems.
Either solving the \gls*{ocp} online to control the robot using \gls*{nmpc} techniques, or offline as a trajectory generator.
In~\cite{lunni_nonlinear_2017}, an \gls*{nmpc} controller at the kinematic level is used to generate set points for a low-level controller in a decentralized architecture.
A closer approach to our current proposal can be found in~\cite{garimella_towards_2015,geisert_trajectory_2016}.
Even though they consider the full-body dynamics, both use \gls*{ocp} to generate a trajectory but not to control the \gls*{am} through \gls*{nmpc}.
Besides, none of them consider contacts.
In~\cite{tzoumanikas_aerial_2020},  a hybrid-\gls*{nmpc} is proposed to perform a writing task with a delta robot.
That is, the \gls*{ocp} in the controller solves the problem at a force level for the platform and at the position level for the end-effector of the robotic arm.
However, the dynamical model is simplified by considering quasi-static maneuvers.

To the best of the authors' knowledge, full-body \gls*{nmpc} at the force/torque level has not been proposed for \glspl*{am}.
However, there have been a few proposals involving single multicopters (not manipulators).
In~\cite{kamel_fastnonlinear_2015},  a \gls*{nmpc} controller is used to track the attitude given by a higher level trajectory generator.
An improved version is shown in~\cite{neunert_fast_2016}, where the planning and the control blocks are unified.
They use a \gls*{slq} solver to solve, online, an \gls*{ocp} and apply the optimized thrust commands to the motors.
A similar approach is taken in~\cite{brunner_trajectory_2020}, where they consider the omnidirectional multicopter described in~\cite{brescianini_omni_2016}.
They use the derivative of the wrench on the body as the control variable in the \gls*{nmpc}, resorting to control allocation to retrieve the rotor speeds and tilt angles.


\section{Optimal Control} \label{sec:optimal_control}

An \gls*{ocp} is a specific type of optimization problem that mainly involves a dynamical system and a cost-function whose value depends on the evolution of this dynamical system.
Its original continuous-time version has the following form:
\begin{equation}
  \begin{aligned}
    & \underset{\mathbf{x}(t), \mathbf{u}(t)}{\min} & & \int_{t=0}^{T}l(\mathbf{x}(t), \mathbf{u}(t)) dt + L(\mathbf{x}(T))\\
    & \text{s.t.} & & \dot{\mathbf{x}} = f_c(\mathbf{x}(t), \mathbf{u}(t)) \,, t\in [0, T] & \text{(dynamics)} \,, \\
    & & & \mathbf{x}_{0} =  \mathbf{x}(0)                                    & \text{(initial value)} \,,        \\
    & & &  h(\mathbf{x}(t), \mathbf{u}(t)) \geq 0                             & \text{(path const.)} \,,     \\
    & & & r(\mathbf{x}(T)) \geq 0                                            & \text{(term. const.)} \,,
  \end{aligned}
  \label{eq:cocp}
\end{equation}
where $\mathbf{x}(t)$ and $\mathbf{u}(t)$ are respectively the state and control trajectories. 
The cost function is composed of the running cost given by $l(\mathbf{x}(t), \mathbf{u}(t))$ and the terminal cost, given by $L(\mathbf{x}(T))$.
The dynamics constraint is represented by an \gls*{ode} containing the model of the system (see \secref{sec:modeling} for details on modeling a \gls*{am}).
Besides, there is also the initial value constraint and, optionally, equality and inequality constraints for the state and the control trajectories.

We find two main approaches for solving the problem~\eqref{eq:cocp},~\cite{betts_practicalmethods_optimalcontrol_book}.
On the one hand, the so-called \emph{indirect methods} are based on finding optimality conditions for the continuous time problem.
Eventually, these conditions are resolved numerically.
Thus, they are also known as \emph{first optimize, then discretize} methods.
We can classify the indirect methods depending on the nature of their solution.
Global indirect approaches are based on the \gls*{hjb} equation, which is derived by applying the dynamic programming principle to the continuous-time \gls*{ocp}.
The big advantage of such methods is that they are able to give global optimal feedback policies.
However, they are hardly applicable to complex systems due to the curse of dimensionality.
This drawback is avoided in local indirect approaches, which apply the \gls*{pmp}, yielding necessary conditions for optimality.
These conditions contain ODEs forming a \gls*{tpbvp} that can be solved numerically, \eg\ using shooting or collocation methods.
It is important to remark that, when solved, the obtained trajectories will only be candidate solutions since \gls*{pmp} only offers necessary conditions for optimality, also known as \emph{First Order Necessary Conditions}.

On the other hand, \emph{direct methods} (also known as \emph{first discretize, then optimize}) transcribe the continuous time formulation (\ie, infinite-variable problem) \eqref{eq:cocp} into a discretized version (\ie, finite-variable) that takes the form of a \gls*{nlp} problem. That is (notice that we omit the path and terminal constraints for the sake of brevity),
\begin{equation}
  \begin{aligned}
    & \underset{\mathbf{X}, \mathbf{U}}{\min} & & \sum_{k=0}^{N-1}l_k(\mathbf{x}_k, \mathbf{u}_k) + L(\mathbf{x}_N) \\
    & \text{s.t.} & & \mathbf{x}_{k+1} = f(\mathbf{x}_k, \mathbf{u}_k) \,, \quad k\in [0, N-1] \,, \\
    & & & \mathbf{x}_{0} =  \mathbf{x}(0) \,,
  \end{aligned}
  \label{eq:docp}
\end{equation}
where $\mathbf{X}=\{\mathbf{x}_0, \dots,\mathbf{x}_N\}$ and $\mathbf{U}=\{\mathbf{u}_0, \dots,\mathbf{u}_{N-1}\}$ indicate, respectively, the set of states and controls that constitute the discrete trajectory.
The dynamic constraint now has become a constraint for the states $\mathbf{x}_k$,\ $\mathbf{x}_{k+1}$ and the control $\mathbf{u}_k$.
The current $f(\mathbf{x}_k, \mathbf{u}_k)$ differs from its continuous time version in the sense that it contains the numerical integration of the \gls*{ode} (see \secref{subsec:modeling_dynamics}).

The discretized problem in~\eqref{eq:docp} can be solved by any usual method aimed at solving constrained \gls*{nlp} problems~\cite{nocedal_numerical_book}.
These methods have become popular thanks to their simple and intuitive formulation.
Besides, they allow the user to focus on such formulation, while leaving the solution to off-the-shelf solvers, \eg\ \textsc{Ipopt}~\cite{wachter_mp_2006} or \textsc{Knitro}~\cite{byrd_knitro_2006}.
These solvers usually rely on methods that perform big matrix factorizations when computing the search direction, something that in general implies to invert the \gls*{kkt} matrix.
This slows down the solving process and, therefore, limits their application either to problems where the solving time is not an issue (\ie, offline \gls*{to}) or \gls*{nmpc} with simple systems that allow an online solution of  \eqref{eq:docp}.

\subsection{Differential Dynamic Programming (DDP)}

\gls*{ddp}~\cite{mayne_secondorder_1966} is a low-complexity algorithm that reduces the computational cost of solving \eqref{eq:docp}, enabling the online solution for complex systems such as an \gls*{am}.
\gls*{ddp} takes advantage of the Markovian character of the \gls*{ocp} and its resulting inherent sparsity: instead of  factorizing one single big matrix, it performs several factorizations of many small matrices.
Though \gls*{ddp} was presented over five decades ago, the method has recently regained attention in the robotics community due to its low computational requirements and also thanks to its new formulations, mainly represented by \gls*{ilqr}~\cite{li_ilqr_2004} and \gls*{slq}~\cite{sideris_slq_2005}.
Using Gauss-Newton approximations, these formulations improve the speed of the original \gls*{ddp} algorithm and are common choices when applying \gls*{nmpc} to complex robotics systems like humanoids or quadrupeds~\cite{neunert_wholebody_2018}.

The \gls*{ddp} method addresses the discretized \gls*{ocp}~\eqref{eq:docp}.
\gls*{ddp} starts with a non-optimal trajectory for the controls $\mathbf{U}\triangleq\mathbf{u}_{0:N-1}$ and states $\mathbf{X}\triangleq\mathbf{x}_{0:N}$ and iteratively computes small improvements. 
This strategy resembles the local indirect methods, which are based on improving an existing trajectory.
However, instead of solving the \gls*{tpbvp} to find an improvement for the trajectory, \gls*{ddp} (like global indirect methods) relies on the dynamic programming technique.
This is why it is normally considered to be an indirect method\footnote{To solve the \gls*{ocp} by means of the \gls*{ddp} algorithm, we need first to discretize the problem. Thus, there exists literature where it is considered to be a direct method, \eg~\cite{koenemann_wholebody_2015}. In fact, if we consider the \gls*{ddp} algorithm as a black-box, it is indeed an \gls*{nlp} solver for problems with a Markovian structure.}. 

Each  iteration of the \gls*{ddp} algorithm contains two principal steps or \emph{passes}, which run over all the nodes (discretization points) of the trajectory: the \emph{backward pass} and the \emph{forward pass}.
In the following we give the main formulation involved in these passes.

\subsubsection{Backward pass}
This pass results in an optimal policy at every node.
We start the derivation of the equations involved in the backward pass by expressing the cost associated to the tail of the trajectory (from any node $i$ until the terminal node $N$), \ie,
\begin{equation} \label{eq:cost_tail}
  J_i(\mathbf{x}_i, \mathbf{U}_i) = \sum_{k = i}^{N - 1}l_k(\mathbf{x}_k, \mathbf{u}_k) + l_N(\mathbf{x}_N) \,.
\end{equation}
Thanks to the Bellman principle, we can recursively solve this problem as
\begin{equation}
  \begin{split}
    V_i(\mathbf{x}_i) &\triangleq \min_{\mathbf{U}_i} J_i(\mathbf{x}, \mathbf{U}_i) \\
    &= \min_{\mathbf{u}_i}[l_i(\mathbf{x}_i, \mathbf{u}_i) + V_{i+1}(f(\mathbf{x}_i, \mathbf{u}_i))]~,
  \end{split}
  \label{eq:optimal_ctg}
\end{equation}
with $V_i(\mathbf{x}_i) \in \mathbb{R}$ being the \emph{optimal cost-to-go} or \emph{value} function.
Then, \gls*{ddp} finds the local optimum of the following expression
\begin{equation}
  \begin{split}
    Q_i(\delta \mathbf{x}_i, \delta \mathbf{u}_i) &= l_i(\mathbf{x}_i + \delta \mathbf{x}_i, \mathbf{u}_i + \delta \mathbf{u}_i)  \\
    &+ V_{i+1}(f(\mathbf{x}_i + \delta \mathbf{x}_i, \mathbf{u}_i + \delta \mathbf{u}_i)) \,.
  \end{split}
  \label{eq:Q_original}
\end{equation}
For the sake of clarity, in the following the sub-indices $i$ are omitted and the optimal \emph{cost-to-go} at the next time step is shown with the prime symbol, \ie, \mbox{$V'(f(\mathbf{x}, \mathbf{u})) \triangleq V_{i+1}(f(\mathbf{x}_i, \mathbf{u}_i)) $.}
The optimum of~\eqref{eq:Q_original} is found by writing the quadratic approximation,
\begin{equation}
  Q(\delta \mathbf{x}, \delta \mathbf{u}) \approx  \cfrac{1}{2}
  \begin{bmatrix}
    1 \\ \delta \mathbf{x} \\  \delta \mathbf{u}
  \end{bmatrix} ^ \top
  \begin{bmatrix}
    0                     & \mathbf{Q}_\mathbf{x}^\top             & \mathbf{Q}_\mathbf{u}^\top        \\
    \mathbf{Q}_\mathbf{x} & \mathbf{Q}_{\mathbf{x}\mathbf{x}}      & \mathbf{Q}_{\mathbf{x}\mathbf{u}} \\
    \mathbf{Q}_\mathbf{u} & \mathbf{Q}_{\mathbf{x}\mathbf{u}}^\top & \mathbf{Q}_{\mathbf{u}\mathbf{u}} \\
  \end{bmatrix}
  \begin{bmatrix}
    1 \\ \delta \mathbf{x} \\  \delta \mathbf{u}
  \end{bmatrix}\,,
  \label{eq:ctg_quadratic}
\end{equation}
and then optimizing with respect to $\delta\mathbf{u}$.
The partial first- and second-order derivatives of~\eqref{eq:Q_original} appearing in \eqref{eq:ctg_quadratic} are,
\begin{subequations}
  \label{eq:Q_partials}
  \begin{align}
    \mathbf{Q}_\mathbf{x}             & = \mathbf{l}_\mathbf{x} + f_\mathbf{x}^\top \mathbf{V}'_{\mathbf{x}} \,,                                                                                                                \\
    \mathbf{Q}_\mathbf{u}             & = \mathbf{l}_\mathbf{u} + f_\mathbf{u}^\top \mathbf{V}'_{\mathbf{x}} \,,                                                                                                                \\
    \mathbf{Q}_{\mathbf{x}\mathbf{x}} & = \mathbf{l}_{\mathbf{x}\mathbf{x}} + f_\mathbf{x}^\top \mathbf{V}'_{\mathbf{x}\mathbf{x}} f_\mathbf{x} + \mathbf{V}'_{\mathbf{x}} \cdot f_{\mathbf{x}\mathbf{x}} \,, \label{subeq:Qxx} \\
    \mathbf{Q}_{\mathbf{u}\mathbf{x}} & = \mathbf{l}_{\mathbf{u}\mathbf{x}} + f_\mathbf{u}^\top \mathbf{V}'_{\mathbf{x}\mathbf{x}} f_\mathbf{x} + \mathbf{V}'_{\mathbf{x}} \cdot f_{\mathbf{u}\mathbf{x}} \,, \label{subeq:Qux} \\
    \mathbf{Q}_{\mathbf{u}\mathbf{u}} & = \mathbf{l}_{\mathbf{u}\mathbf{u}} + f_\mathbf{u}^\top \mathbf{V}'_{\mathbf{x}\mathbf{x}} f_\mathbf{u} + \mathbf{V}'_{\mathbf{x}} \cdot f_{\mathbf{u}\mathbf{u}} \,, \label{subeq:Quu}
  \end{align}
\end{subequations}
where $\mathbf{V}'_{\mathbf{x}}$,~$(\mathbf{l_x},~\mathbf{l_u})$,~$(\mathbf{f_x,~f_u})$ and $\mathbf{V}'_{\mathbf{x}\mathbf{x}}$,~$(\mathbf{l_{xx},~l_{ux},~l_{uu}})$, $(\mathbf{f_{xx},~f_{ux},~f_{uu}})$ describe the Jacobians and Hessians of the value, cost and dynamics functions, respectively\footnote{%
Note that the last terms of \eqref{subeq:Qxx}-\eqref{subeq:Quu}, denoting the product of a vector by a tensor, are not considered in the \gls*{ilqr} approach.}.

The minimization of~\eqref{eq:ctg_quadratic} with respect to $\delta \mathbf{u}$ leads to the  optimal policy
\begin{equation}
  \delta \mathbf{u}^* (\delta \mathbf{x}) = \mathbf{k} + \mathbf{K}\delta \mathbf{x} \,,
  \label{eq:optimal_policy_ddp}
\end{equation}
with $\mathbf{k} \triangleq  -\mathbf{Q}_{\mathbf{u}\mathbf{u}}^{-1}\mathbf{Q}_\mathbf{u}$ and $\mathbf{K} \triangleq -\mathbf{Q}_{\mathbf{u}\mathbf{u}}^{-1}\mathbf{Q}_{\mathbf{u}\mathbf{x}}$.
If we insert this optimal policy into the quadratic expansion in \eqref{eq:ctg_quadratic}, we obtain a quadratic approximation of the optimal \emph{cost-to-go} as a function of $\delta \mathbf{x}$, \ie
\begin{equation}
  V(\delta \mathbf{x}) = \Delta V + \mathbf{V}_{\mathbf{x}}^\top \delta \mathbf{x} + \delta\mathbf{x}^\top \mathbf{V}_{\mathbf{x}\mathbf{x}}\delta\mathbf{x} \,,
  \label{eq:V_improvement}
\end{equation}
where
\begin{subequations}
  \label{eq:Value_update}
  \begin{align}
    \Delta V &= -\cfrac{1}{2}\mathbf{k}^\top \mathbf{Q}_{\mathbf{u} \mathbf{u}} \mathbf{k} \label{eq:single_expected_reduction}\,,\\
    \mathbf{V}_{\mathbf{x}} &= \mathbf{Q}_\mathbf{x} - \mathbf{K}^\top \mathbf{Q}_{\mathbf{u}\mathbf{u}} \mathbf{k} \,,\\
    \mathbf{V}_{\mathbf{x}\mathbf{x}} &= \mathbf{Q}_{\mathbf{x}\mathbf{x}} - \mathbf{K}^\top \mathbf{Q}_{\mathbf{u}\mathbf{u}}\mathbf{K} \,.
  \end{align}
\end{subequations}
We start the backward pass by assigning $V_N(\mathbf{x}_N) = L(\mathbf{x}_N)$ and then using the set of equations~\eqref{eq:Q_partials} to~\eqref{eq:Value_update} to update recursively the optimal policy at each node.

Notice finally that at the optimum we will have $\bf Q_u=0$ and thus the optimal policy \eqref{eq:optimal_policy_ddp} becomes 
\begin{align}\label{eq:optimal_policy}
\delta\mathbf{u}=\bf K\delta x~.
\end{align}
This optimal policy can be used  outside the \gls{ddp} algorithm as a state feedback law to modify the control commands in case of observed variations of the state $\bf x$. 

\subsubsection{Forward pass}
Given a current guess of the trajectories $(\mathbf{X,U})$ and the optimal policies computed during the backward pass, the forward pass applies appropriate modifications to them, node by node and proceeding forward,
\begin{subequations}
\label{eq:forward_pass}
  \begin{align}
    \hat{\mathbf{x}}_0 &= \mathbf{x}_0 \,, \\
    \hat{\mathbf{u}}_k &= \mathbf{u}_0 + \alpha \mathbf{k}_k + \mathbf{K}_k(\hat{\mathbf{x}}_k - \mathbf{x}_k)\,,  \label{subeq:ddp_control_update}\\
    \hat{\mathbf{x}}_{k+1} &\triangleq f(\hat{\mathbf{x}}_k, \hat{\mathbf{u}}_k) \label{subeq:rollout}\,,
  \end{align}
\end{subequations}
where the $\hat{\mathbf{x}}$ and $\hat{\mathbf{u}}$ are respectively the updated values for the states and controls.
The assignment in~\eqref{subeq:rollout} implies the integration of the dynamics to update the next state.
This procedure is also known as the \emph{non-linear rollout}.
The parameter $\alpha\in(0,1]$ indicates the length of the step taken by the current iteration.
A value of $\alpha=1$ results in the application of a full step.

\subsubsection{Improvements}
In its original form, \gls*{ddp} has poor globalization capabilities, \ie, from an arbitrary initial guess it struggles to converge to a good optimum.
Mainly, this is due to the fact that \gls*{ddp} does not allow for infeasible trajectories during the optimization process, \ie, the constraints \mbox{$\mathbf{x}_{k+1} = f(\mathbf{x}_k, \mathbf{u}_k)$} have to be always fulfilled, like in a single shooting approach (see~\cite{betts_practicalmethods_optimalcontrol_book}).
There have been several works trying to emulate a multiple shooting algorithm with \gls*{ddp} like \eg~\cite{giftthaler_family_2018} or~\cite{mastalli_crocoddyl_2020} where they present a solver called \gls*{fddp}, whose numerical behavior resembles the classical multiple shooting approach. \gls*{fddp} considerably improves the globalization capability and the convergence rate of the solver.

Another limitation of the original \gls*{ddp} is its inability to handle constraints other than the ones imposed by the system dynamics. 
This issue has been tackled differently in several works, \eg~\cite{howell_altro_2019,mastalli_boxfddp_2020,kazdadi_eqddp_2021}.

In this paper, we use an improved version of the \gls*{fddp} algorithm~\cite{mastalli_crocoddyl_2020} that bounds the control inputs  by means of a squashing function~\cite{marti_squash-box_2020}.
Other constraints are modeled as soft-constraints, that is, through quadratic barriers in the cost function.


\section{Modeling} \label{sec:modeling}

\subsection{Dynamics}
\label{subsec:modeling_dynamics}
The discrete-time evolution of the dynamical system from time $t_k$ to time $t_{k+1}$ has been defined in \eqref{eq:docp} as
\begin{equation}
  \label{eq:discrete_system_dynamics}
  \mathbf{x}_{k+1} = f(\mathbf{x}_k, \mathbf{u}_k) \,.
\end{equation}
This expression is used as a short-hand version of the true equation, defined with
\begin{equation}
  \label{eq:state_integration}
  \mathbf{x}_{k+1} = \mathbf{x}_{k} \oplus  \int_{t_k}^{t_{k+1}} \dot{\mathbf{x}}(t) \,dt \,,
\end{equation}
where the $\oplus$ symbol stands for the right-plus operation on the manifold to which $\mathbf{x}$ belongs~\cite{sola_micro_2018}, \ie,  $\mathbf{x}\oplus \mathbf{v} \triangleq \mathbf{x}\cdot\exp(\mathbf{v})$ for elements on non-linear manifolds such as the $SE(3)$ platform pose, and $\bf u\oplus v \triangleq u + v$ for regular vectors such as velocities and joint angles.
Evaluating \eqref{eq:state_integration} involves two main steps: $i)$ the computation of the state derivative according to the continuous-time dynamical model of the platform, $\dot{\mathbf{x}}(t) = f_c(\mathbf{x}(t), \mathbf{u}(t))$; and $ii)$ its numerical integration over the time step $\Delta t = t_{k+1} - t_{k}$.
In this section we focus on specifying the model for the aerial manipulator $(i)$, \ie, on finding $\dot{\mathbf{x}}(t) = f_c(\mathbf{x}(t), \mathbf{u}(t))$.
The integration is performed  numerically, and the  method used should ensure a good balance between accuracy and computational cost~\cite{rawlings_modelpredictivecontrol_book}.
We use for this either the semi-implicit Euler or the Runge-Kutta RK4 methods, the latter being more costly but more accurate.

\begin{figure}[tb]
  \centering
  \includegraphics[trim=0em 10.0em 10em 0.0em,clip,width=0.65\linewidth]{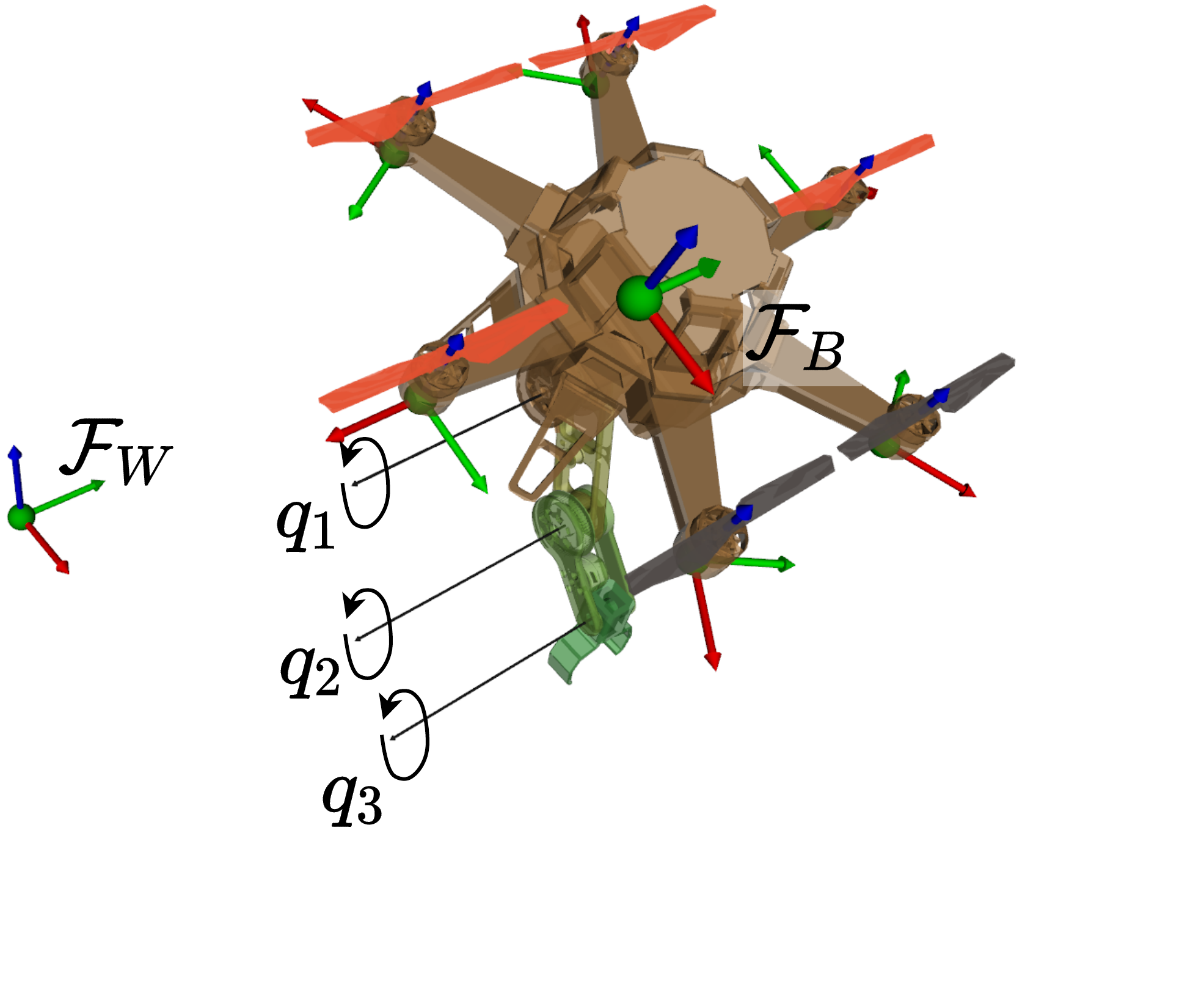}
  \caption{Frame definitions for a typical \gls*{am} consisting of a planar hexacopter with a 3-\gls*{dof} serial robot arm. $\mathbf{q}_{B}$ represents the rigid transformation between $\mathcal{F}_{W}$ and $\mathcal{F}_{B}$. In this case, \mbox{$\mathbf{q}_J = [q_1, q_2, q_3]^\top$}.}
  \label{fig:frames}
\end{figure}

In the following, we express the pose of the flying platform or \emph{base link} as an element $\mathbf{q}_{B} \in SE(3)$, and the configuration of the arm as a vector of the $n_J$ joint angles $\mathbf{q}_{J} \in \mathbb{R}^{n_J}$ (\figref{fig:frames}). 
We note the robot's configuration as $\mathbf{q} = (\mathbf{q}_{B}, \mathbf{q}_{J})$.
Since we are interested in the dynamics, the robot state $\bf x$ contains the  configuration and its time derivative, $\mathbf{x} = \{\mathbf{q}, \dot{\mathbf{q}}\}$.
The derivative of $SE(3)$ is a vector of linear and angular velocities in its tangent space $se(3)$, isomorph to $\mathbb{R}^6$, so that $\dot{\mathbf{q}}_{B}=(\mathbf{v},\bm{\omega})\in \mathbb{R}^6$ and $\dot{\mathbf{q}}\in \mathbb{R}^{6+n_J}$.
The state derivative $\mathbf{\dot{x}}$ can thus be represented by a regular vector $\mathbf{\dot{x}} = [\dot{\mathbf{q}}^\top, \ddot{\mathbf{q}}^\top]^\top \in \mathbb{R}^{12+2n_J}$.

The expression for $\dot{\mathbf{x}} = f_c(\mathbf{x}, \mathbf{u})$ can be written as,
\begin{align}
\mathbf{\dot x} =
    \begin{bmatrix}
      \mathbf{\dot q}\\\mathbf{\ddot q}
    \end{bmatrix}
    =
    \begin{bmatrix}
      \mathbf{\dot q}\\
      \textit{FD}(\mathbf{q},\mathbf{\dot q},\mathbf{u})
    \end{bmatrix}~,
\end{align}
of which we only need to find the accelerations $\ddot{\mathbf{q}}$.
These result from the effect of all the forces applied to the robot, and can be found by computing the \gls*{fd}.
Here, we distinguish among two cases depending on whether or not the system's motion is constrained by external contacts.
Both are detailed hereafter.

\subsubsection{Free flying phase}
\label{subsubsec:free_flying}
When the robot is not subject to any contact constraining the motion, we can express the system equation of motion as~\cite{featherstone_rigidbody_book}
\begin{equation}
  \mathbf{H}(\mathbf{q})\mathbf{\ddot{q}} + \mathbf{C(\mathbf{q}, \dot{\mathbf{q}})} = \bm{\tau}(\mathbf{u}) \,,
  \label{eq:motion_free_flying}
\end{equation}
where $\mathbf{H}$ is the generalized inertia matrix, $\mathbf{C}$ is a force vector containing the Coriolis, centrifugal and gravitational terms and $\bm\tau(\mathbf{u})$ is the joint generalized torque produced by the inputs $\mathbf{u}$ (see Section~\ref{sec:actuation_model}).
To compute the \gls*{fd} (\ie, to isolate $\mathbf{\ddot{q}}$ from \eqref{eq:motion_free_flying}) we do not necessarily need to compute $\mathbf{H}^{-1}$, which is  expensive specially for complex robots.
It is much more convenient to resort to the  \gls*{aba}, explained in \cite{featherstone_rigidbody_book}, which significantly reduces the cost of computing the \gls*{fd} thanks to its recursive nature.

\subsubsection{Contact phase}
\label{subsubsec:contact}
Any contact of a robot link with the environment constrains the robot's motion.
In this paper, we account for rigid contacts, modeling them as implicit holonomic constraints, \ie,
\begin{equation}
  \phi(\mathbf{q}) = 0 \,.
  \label{eq:contact_pos_constraint}
\end{equation}
For example, a punctual contact of a link, with $\mathbf{p}_b(\mathbf{q})$ the point of contact  belonging to the link, constrains the linear \glspl*{dof} of the body at the environment's contact point $\mathbf{p}_c$, such as \mbox{$\phi(\mathbf{q}) = \mathbf{p}_b(\mathbf{q}) - \mathbf{p}_c = 0$}.
As the contact in~\eqref{eq:contact_pos_constraint} is considered static, its time derivatives should also be null, \ie,%
\begin{subequations}
  \begin{align}
    \label{eq:contact_vel_constraint}
    \dot{\phi}(\mathbf{q})  & = \mathbf{J}_c \dot{\mathbf{q}} = 0 \,,                                        \\
    \label{eq:contact_acc_constraint}
    \ddot{\phi}(\mathbf{q}) & = \mathbf{J}_c \ddot{\mathbf{q}} +  \dot{\mathbf{J}}_c \dot{\mathbf{q}} = 0 \,,%
  \end{align}%
  \label{eq:conatct_constraints}%
\end{subequations}%
where $\mathbf{J}_c = \frac{\partial \phi}{\partial \mathbf{q}}$ is the contact Jacobian matrix, which maps the joint velocities to velocities at the contact point.

Thanks to the duality between velocity and force, this Jacobian's transpose $\mathbf{J}_c^\top$ also maps each individual contact force to the joint-space~\cite{featherstone_rigidbody_book}. 
These external forces, represented by $\bm{\lambda}$, affect the evolution of the system, so we add them to \eqref{eq:motion_free_flying}, yielding the new dynamics equation,
\begin{equation}
  \mathbf{H}(\mathbf{q})\mathbf{\ddot{q}} + \mathbf{C(\mathbf{q}, \dot{\mathbf{q}})} = \bm{\tau}(\mathbf{u}) + \mathbf{J}_c^\top \bm{\lambda} \,.
  \label{eq:motion_contact}
\end{equation}
In order to solve this new \gls*{fd} problem we need to find the external forces $\bm{\lambda}$.
This must be done subject to the constraint~\eqref{eq:contact_pos_constraint} and its time derivatives~\eqref{eq:conatct_constraints}, which guarantee that the contact will remain static.
Imposing that we reach the contact at zero velocity, that is without impact, we can ensure smooth trajectories for all variables (otherwise \eqref{eq:motion_contact} would become singular at the contact instant). 
Then, we can find the joint acceleration and the contact forces by solving the system  formed by equations \eqref{eq:motion_contact} and \eqref{eq:contact_acc_constraint}, which can be posed in compact form as,
\begin{equation}
  \begin{bmatrix}
    \mathbf{H} & \mathbf{J}_c^\top \\
    \mathbf{J}_c & \mathbf{0}
  \end{bmatrix}
  \begin{bmatrix}
    \ddot{\mathbf{q}} \\ - \bm{\lambda}
  \end{bmatrix}
  = 
  \begin{bmatrix}
    \bm{\tau} - \mathbf{C(\mathbf{q}, \dot{\mathbf{q}})} \\
    -\dot{\mathbf{J}}_c \dot{\mathbf{q}}
  \end{bmatrix} \,.
  \label{eq:contact_kkt}
\end{equation}
Unfortunately, this form is not suited for the \gls*{aba} algorithm, and the solving methods often resort to factoring the \gls*{kkt} matrix through Cholesky decomposition---see~\cite{featherstone_rigidbody_book}. 

An alternative \gls*{fd} computation, explicitly solving only for $\ddot{\mathbf{q}}$, consists in applying the Gauss principle of least constraint~\cite{featherstone_rigidbody_book,budhiraja_ddp_multiphase_2018}.
According to this, the resulting joint acceleration will be the closest one to a free motion acceleration, \ie, the eventual acceleration of the system in the absence of contacts.
This can be posed as the \gls*{qp} problem
\begin{equation}
  \begin{aligned}
    &\underset{\ddot{\mathbf{q}}}{\min} & & \cfrac{1}{2} \| \ddot{\mathbf{q}} - \ddot{\mathbf{q}}_{\text{free}} \|^2_{\mathbf{H}(\mathbf{q})} \\
    & \text{s.t.} & & \mathbf{J}_c(\mathbf{q}) \ddot{\mathbf{q}} + \dot{\mathbf{J}}_c(\mathbf{q}) \dot{\mathbf{q}} = 0 \,.
  \end{aligned}
  \label{eq:gauss_principle}
\end{equation}
Interestingly, the \gls*{kkt} conditions of optimality for \eqref{eq:gauss_principle} are given by \eqref{eq:contact_kkt}, and so the Lagrange multipliers for solving \eqref{eq:gauss_principle} are precisely the vector $\bm{\lambda}$, which is implicitly found alongside $\mathbf{\ddot q}$. 
This renders the problems $\eqref{eq:contact_kkt}$ and $\eqref{eq:gauss_principle}$ equivalent.
In other words, while $\bm{\lambda}$ represents the dual variables (or Lagrange multipliers) from an optimization perspective, it represents the contact forces from the mechanical perspective.

\subsection{Actuation model}
\label{sec:actuation_model}

We express the relation between control and generalized force by means of a linear mapping
\begin{equation}
  \mathbf{\bm\tau}(\mathbf{u}) = \mathbf{G}\mathbf{u} \,.
\end{equation}
The mapping matrix $\mathbf{G}$ has to be properly set according to the robot's geometry.
For instance, in a fully-actuated robot all \glspl*{dof} are actuated and so $\mathbf{G}$ is full-rank.
However, this is not the case for most \gls*{am} since they are built upon planar multicopter platforms. 
To obtain an expression for $\mathbf{G}$, let us express $\mathbf{G}$ as a block-diagonal matrix containing $\mathbf{G}_{B}$ and $\mathbf{G}_{arm}$, with the latter mapping the arm joints' torques.
In this work, we consider the arms are fully-actuated, \ie,  $\mathbf{G}_{arm} = \mathbf{I}_{n_J}$.
On the other hand, the matrix $\mathbf{G}_{B}$ maps controls to the generalized force applied to the flying platform.
Hence, $\mathbf{G}_{B}$ synthesizes information about the geometry of the flying platform, \ie, the number of propellers, how they are distributed and how their axes of rotation are placed.
All these factors will make a significant impact on the behavior of the \gls*{am}.
For example, the platform can be under-actuated if all the axes of the propellers are perpendicular to the platform.
Contrarily, they can be strategically tilted so that the platform becomes fully-actuated.
The remaining of this section is devoted to properly define $\mathbf{G}_{B}$.

Let us consider a reference frame $\{^{B}\mathbf{p}_i,\mathbf{R}_{B, i}\}$ for each propeller $i$, expressed in the base link reference, whose z axis is aligned with the propeller's rotation axis (see \figref{fig:frames}).
Let $u_i$ be the thrust generated by the $i$-th propeller and used as control input to the system (the thrust of a propeller is well approximated by the square of its rotational velocity, $u_i=c_f\, \omega_i^2$).
This thrust produces an external force to the multirotor platform, which has the expression
\begin{equation}
  ^{B}\mathbf{f}_{i, th} = u_i \mathbf{R}_{B, i} \mathbf{e}_3 \,,
\end{equation}
where $\mathbf{e}_3$ is the standard unit vector along the z direction.
For the same rotor $i$, the resulting couple $^{B}\mathbf{n}_{B, i}$ applied to the platform  comes from the displaced thrust and from the drag moment, that is, 
\begin{equation}
  ^{B}\mathbf{n}_{B, i} = {}^{B}\mathbf{p}_i \times \mathbf{f}_{i, th} + (-1)^{CCW} u_i \cfrac{c_m}{c_f}\mathbf{R}_{B, i} \mathbf{e}_3 \,,
\end{equation}
where $c_m$ and $c_f$ are constants related to the propeller's geometry.
If the spinning rotation of the $i$-th rotor is \emph{counter-clockwise}, $(-1)^{CCW}$ evaluates to $-1$, and to $1$ otherwise.
Stacking both the force and the couple for the $i$-th propeller, we get the $i$-th column for the matrix $\mathbf{G}_{B}$, \ie,
\begin{align}
  \begin{split}
    \bm\tau_{B,i} &= \mathbf{g}_i u_i  \\
    &=
    \begin{bmatrix}
      \mathbf{R}_{B, i} \mathbf{e}_3 \\
      ^{B}\mathbf{p}_i \times \mathbf{R}_{B, i} \mathbf{e}_3 + (-1)^i \cfrac{c_m}{c_f}\mathbf{R}_{B, i} \mathbf{e}_3
    \end{bmatrix}
    u_i \,.
  \end{split}
\end{align}
The matrix $\mathbf{G}_{B}$ results from writing the above for all propellers in matrix form, 
\begin{equation}
  \bm\tau_{B} =
  \begin{bmatrix}
    \mathbf{g}_{1} & \cdots & \mathbf{g}_{n_{prop}}
  \end{bmatrix}
  \begin{bmatrix}
    u_1 \\ \vdots \\ u_{n_{prop}}
  \end{bmatrix}
  \triangleq \mathbf{G}_{B} \mathbf{u}_{prop} \,.
\end{equation}


\section{Cost Function} 
\label{sec:cost_functions}

The two essential parts of an \gls*{ocp} are the system dynamic constraints, which approximate the behavior of the real \gls*{am} (see \secref{sec:modeling}), and the cost function, which specifies the optimality criteria for fulfilling the tasks.
The latter represents  an intuitive manner to specify control objectives.
A proper engineering of the cost function and a good weight tuning are crucial to obtain the desired trajectories that drive the \gls*{am} to fulfill required tasks.
In this section we specify the main structure of a cost function along a trajectory and describe some typical residuals (costs) that are useful when applying optimal control to aerial manipulation.

\subsection{Structure}
\label{subsec:cf_structure}

\begin{figure}[t]
  \centering
    \includegraphics[trim=10 80 0 10, clip, width=0.9\linewidth]{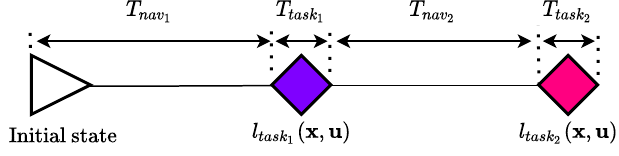}
    \vspace{-0.5em}
    \includegraphics[trim=10 0 0 100, clip, width=0.9\linewidth]{figures/oc_trajectory_specification.pdf}
  \caption{Top: trajectory specification considering two tasks and two navigation phases. Bottom: a possible translation to an \gls*{ocp}, containing a series of nodes uniformly distributed in time.}
  \label{fig:trajectory_specification}
\end{figure}

We structure missions by a concatenation of movements or \emph{phases}, which we divide in two main types: task  phases and navigation phases (see \figref{fig:trajectory_specification}).
Task phases correspond to the periods of time where the robot is required to do something specific or to be somewhere. 
Waypoints, contacts or manipulations are considered tasks. 
They are typically implemented through quadratic costs on (parts of) the state, or on functions of the state.
Navigation phases correspond to the periods of time when the robot flies freely from a finished task to the next.
They are mainly constrained by the robot dynamics model.
Navigation often requires the execution of maneuvers that exploit the full-body dynamics of the robot.
This is especially true when dealing with under-actuated platforms.\footnote{A simple toy example is the swing up maneuver for an acrobot, which is a double pendulum only actuated at the elbow. In order to reach the unstable equilibrium state from the stable equilibrium point, we need to plan a complex trajectory that uses the arm dynamics to put both links upright.}
These maneuvers are discovered by the optimization process, provided that the time allocated to the navigation phase is sufficient.

In this work, the timing for each phase will be determined beforehand.
For each phase we shall specify its duration $T_{ph}$ and the structure of the residual (cost).
Given this timeline, optimization will be performed with respect to states and control, discretized over a series of nodes with a uniform sampling time $\Delta t$ (Fig.~\ref{fig:trajectory_specification}).
Given a phase duration $T_{ph}$, the phase's  number of nodes in the \gls*{ocp} is given by $N_{ph} = \frac{T_{ph}}{\Delta t}  + 1$.
Generally the costs are written as the squared weighted 2-norms of different residuals, yielding least-squares formulations which are easy to handle.
Thus, the cost function for the $k$-th node related to a specific phase with $R_{ph}$ associated residuals is
\begin{equation}
  \displaystyle
l_{k,ph}(\mathbf{x}_k, \mathbf{u}_k) =\sum_{i=1}^{R_{ph}} w_i \|\mathbf{r}_i(\mathbf{x}_k, \mathbf{u}_k)\|^2_{\mathbf{W}_i} \,,
\label{eq:to_phase_cost_function}
\end{equation}
where $\mathbf{r}_i(\mathbf{x}_k, \mathbf{u}_k)$ is a vector-valued residual that specifies the $i$-th task for the $k$-th node  (see below for residuals specifications), $\mathbf{W}_i$ is a square weighting matrix, usually diagonal, $\|\mathbf{v}\|^2_{\mathbf{W}}\triangleq \mathbf{v}^\top\mathbf{W}^{-1}\mathbf{v}\in\mathbb{R}$ is the Mahalanobis norm squared and $w_i$ is a scalar to weight the overall task cost.

Navigation phases often contain costs devoted to minimize the energy of the system or the control inputs, \ie, regularization terms for the state and the controls. 
Instead, the task phases consider costs related to the task, for instance the deviation between the desired and current poses of a link.
For this reason, the weights $w_i$ associated to navigation costs are typically (much) smaller than those of the task costs.
This lets the optimizer freely find the best navigation path while strongly enforcing the fulfillment of  the tasks.

\subsection{Typical residuals}

In the following, we present a set of residuals useful for \gls*{ocp} tasks involving \glspl*{am}.
These will be used in the experiments shown in \secref{sec:validation}.
\label{subsec:residuals}

\subsubsection{State error}

A common manner to specify a task is by setting a desired state $\mathbf{x}^*$ for a specific node.
Thus, we can define a state residual as
\begin{align}
    \mathbf{r}_{\text{state}, k} &=  \mathbf{x}^* \ominus \mathbf{x}_k  
    = \begin{bmatrix}
    \text{Log}(\mathbf{M}_{B,k}^{-1}\mathbf{M}_{B}^*) \\ 
    \mathbf{q}^*_J - \mathbf{q}_{J,k} \\
    \dot{\mathbf{q}}^* - \dot{\mathbf{q}}_k \end{bmatrix}
    \,.
  \label{eq:res_state}
\end{align}
The operator $\ominus$ stands for the difference between states expressed as a vector of the space tangent to the states' manifold (see~\cite{sola_micro_2018}): for elements in non-linear manifolds we have $\mathbf{Y} \ominus \mathbf{X}\triangleq\text{Log}(\mathbf{X}^{-1}\mathbf{Y})$, and in vector spaces simply $\mathbf{y}\ominus\mathbf{x}\triangleq\mathbf{y}-\mathbf{x}$.
With this in mind, we can very well define residuals involving only a  part of the state, \eg,
\begin{align*}
\mathbf{r}_{\text{pos},k} &= \mathbf{p}^* - \mathbf{p}_k && \text{(position)} \\
\mathbf{r}_{\text{ori},k} &= \mathbf{R}^* \ominus \mathbf{R}_k && \text{(orientation)} \\
\mathbf{r}_{\text{linv},k} &= \mathbf{v}^* - \mathbf{v}_k && \text{(lin. vel.)} \\
\mathbf{r}_{\text{angv},k} &= \bm{\omega}^* - \bm{\omega}_k && \text{(ang. vel.)} 
~.
\end{align*}

\subsubsection{Control}
Similarly to the state residual, we can define a control residual as
\begin{equation}
  \mathbf{r}_{\text{control}, k} =  \mathbf{u}^* - \mathbf{u}_k \,.
  \label{eq:res_control}
\end{equation}
Normally, this residual is used as a regularization cost, considering $\mathbf{u}^* = \mathbf{0}$ and giving it a low weight.

\subsubsection{Pose and velocity of frames}
It is a common practice in robotics to add a desired task in the operational space, \eg\ a desired pose for the end-effector.
It might also be useful to define desired poses for any arbitrary frame $\mathcal{F}$ in the robot. Thus,
\begin{equation}
  \mathbf{r}_{\text{pose}, k} = \mathbf{M}_{\mathcal{F}}^* \ominus \mathbf{M}_{\mathcal{F},k} = \text{Log}(\mathbf{M}_{\mathcal{F},k}^{-1}\mathbf{M}_{\mathcal{F}}^*) \,,
  \label{eq:res_pose}
\end{equation}
where $\mathbf{M}_{\mathcal{F}}^* \text{, } \mathbf{M}_{\mathcal{F},k} \in SE(3)$ are the desired pose and the pose of the frame $\mathcal{F}$ at the node $k$, respectively.
These frames and their velocities can be obtained from the state, following the method in~\cite{featherstone_rigidbody_book}, by properly concatenating several frame transformations.
%
Therefore, we can also specify a desired linear and/or angular velocity for a frame $\mathcal{F}$ at node $k$, \ie,
\begin{equation}
  \mathbf{r}_{\text{velocity}, k} = \mathbf{v}^*_{\mathcal{F}} - \mathbf{v}_{\mathcal{F}, k} \,.
  \label{eq:res_velocity}
\end{equation}
As before, we can define residuals involving only parts (position, orientation, etc.) of the frame.

\subsubsection{Contact cone}

When defining the contact model in \secref{subsubsec:contact} we assumed null velocity at the contact point.
However, depending on the direction of the contact force and the surface characteristics, slippages may occur.
To avoid them, we constrain the contact forces to be inside the Coulomb's dry friction cone defined by the friction coefficient $\mu$ between the contact point and the surface.
For computation purposes, we approximate the friction cone by a square pyramid, \ie, we impose the conditions
\begin{subequations}
  \label{eq:contact_cone}
  \begin{align}
  \left|f_{c,x} \right| &\leq \mu f_{c,z}  \,, \\ 
  \left|f_{c,y} \right| &\leq \mu f_{c,z}  \,, \\ 
  f_{c,z} &> 0 \,,
  \end{align}
\end{subequations}
where $\mathbf{f}_c = [f_{c,x}, f_{c,y}, f_{c,z}]^\top$ is the contact force represented by $\bm{\lambda}$ in \eqref{eq:motion_contact} expressed in the reference frame of the contact surface.
Notice that we can express \eqref{eq:contact_cone} in matrix form,
\begin{equation}
  \mathbf{A}\mathbf{f}_c  = 
  \begin{bmatrix*}[r]
    1 & 0 & -\mu \\
    -1 & 0 & -\mu \\
    0 & 1 & -\mu \\
    0 & -1 & -\mu \\
    0 & 0 & -1
  \end{bmatrix*}
  \mathbf{f}_c \leq \mathbf{0} \,,
\end{equation}
where we use component-wise inequality.
In this paper we are considering a \gls*{ddp} solver that cannot handle this kind of hard constraints. We therefore add them as soft constraints in the cost function by means of a quadratic barrier, \ie, through a residual whose five components are assigned according to 
\begin{equation}
  r^i_{\text{contact}, k} =
  \begin{cases}
    0 & \textrm{if~~} \mathbf{a}^i\mathbf{f}_c \leq 0 \,, \\ 
    \mathbf{a}^i\mathbf{f}_c & \textrm{if~~} \mathbf{a}^i\mathbf{f}_c > 0 \,,
  \end{cases}
  \text{ for } i=1,\dots,5 \,,
  \label{eq:res_contact_cone}
\end{equation}
where $\mathbf{a}^i$ represents the $i$-th row of the matrix $\mathbf{A}$ above.


\section{Non-linear model predictive control} \label{sec:mpc}

Non-linear model predictive control (nMPC), also referred to as receding horizon control, is a closed-loop control strategy for non-linear systems in which the applied input is determined online by solving an open-loop optimal control problem (\gls{ocp}, \secref{sec:optimal_control}) over a fixed prediction horizon. 

\subsection{Architecture}
\label{subsec:architecture}

We designed an \gls*{nmpc} pipeline  where each time step is triggered by the arrival of robot states, which are provided by some state estimation module.
At every time step, the \gls*{ocp}'s initial state is set with the new robot state, the cost function is updated, the \gls*{ocp} is solved and finally the first command is applied.
This section covers important considerations to achieve a good real-time flow of all these operations.

By construction, the complexity of the \gls*{fddp} algorithm used to solve the \gls{ocp} online is linear with the number of nodes, $N$, cubic with the number of control variables, and linear with the number of iterations, $n$.
It is a common practice to fix the number of iterations to a small value  $n\gtrsim 1$ to keep the overall execution time under control.
While this might terminate the \gls{ocp} before full convergence, the control to apply is the oldest in the \gls{ocp} and has already undergone $N\times n$ iterations, which is enough. 
With $n$ fixed, and since we cannot reduce the number of decision variables (it is robot-dependent), the computing time  is determined by the {number of nodes}.
Given a {node period} $\Dt$, the {length of the receding horizon} of the \gls*{nmpc} is simply $T_H = (N-1)\Delta t$. 
Having a controller with a decent prediction capability $T_H$ is fundamental to fully exploit the possibilities of an under-actuated system,
because these depend on the ability of the \gls{ocp} to discover non-trivial maneuvers.
$T_H$ can be made longer by increasing $\Delta t$, which in turn allocates more computing time, thus allowing for a higher $N$. 
Unfortunately, increasing $\Dt$ augments the granularity of the control, and also compromises the validity of the integration methods (\secref{sec:modeling}), rendering the results inaccurate and/or the system unstable.
It is therefore imperative to find a good balance between the number of nodes $N$, the prediction horizon $T_H$ and the node period $\Dt$.

The horizon $T_H$ is typically shorter than the duration of the mission, $T_M$.
This means that while most nodes of the \gls{ocp} can be warm-started with the previous solution, the terminal (or horizon) node has to incorporate a new setpoint that is just beyond the previous horizon.
These setpoints (\ie, the residuals and the warm-start values) need to be determined based on mission information through some kind of mission plan. 
Since the granularity of the \gls{ocp} is $\Dt$, we need to bring the mission plan, which might be very sparsely specified, to something denser in time.
An obvious strategy is to solve offline an \gls{ocp} problem for the whole mission, and use it as a reference trajectory $\mathbf{X}^*=\{\mathbf{x}_0^*,\cdots,\mathbf{x}_M^*\}$ to progressively feed the online \gls{ocp} of the controller.
For simplicity, it would be desirable to have the time step of the reference trajectory,  $\Dt^*$, equal to the \gls{ocp}'s $\Dt$. 
However, we found that we require $\Dt^*<\Dt$ because the full mission \gls{ocp}, which is much longer and usually not warm-started, is prone to instability and local minima when using large integration node periods.
Due to this step difference, to feed this reference trajectory into the online \gls{ocp} we have to interpolate the reference states $\mathbf{x}_{k-}^*$ and $\mathbf{x}_{k+}^*$ around the \gls{ocp}'s node $k$, 
\begin{align}
    \mathbf{x}_k &=\mathbf{x}_{k-}^* \oplus s\, (\mathbf{x}_{k+}^* \ominus\mathbf{x}_{k-}^*) 
\,,
  \label{eq:slerp}
\end{align}
where $s=(t_k-t_{k-})/\Dt^*\in[0,1]$ is the interpolation factor and $\oplus,\ominus$ have been defined in Sections \ref{sec:modeling} and \ref{sec:cost_functions}.
Alternatively, we can also use \gls*{fd} to integrate the controls over $\mathbf{x}_{k-}^*$ from $t_{k-}$ to $t_k$, which is more accurate but also more expensive.
Finally, notice that this larger \gls{ocp} is slower to solve than the \gls*{nmpc} rate but, since we also use \gls*{fddp} and we only have to solve it once, it is fast enough for a real-time operation.
In case of a very large mission, we can break it into smaller overlapping sub-missions covering  a few \gls{nmpc} horizons each. 
We can then  solve them online in a separate thread and have always references ready to feed the controller.

When the horizon reaches and then surpasses the end of mission at time  $T_M$, we fill the \gls{ocp} nodes beyond $T_M$ with task nodes having the same terminal setpoint at $T_M$, that is $\mathbf{x}_{t\ge T_M}=\mathbf{x}_M^+$.
This setpoint is required to be a statically stable state for the UAM -- \eg\ a hovering state or a landed state. 

A common issue of \gls*{nmpc} controllers is the delay in the control command due to the computation time $T_c$ of the solver.
Given a state estimate $\hat {\bf x}(t_k)$ at time $t_k$, the optimal control to apply will only be known at $t_k+T_c$, and applying it at this late time would be suboptimal~\cite{diehl_fastdirect_2006}. 
Having a predefined number of iterations allows us to anticipate this computation time.
Thus, instead of solving the \gls{ocp} with the first node at $\hat {\bf x}(t_k)$, we use a state $\hat {\bf x}(t_k+T_c)$ predicted using our dynamical model~\eqref{eq:state_integration}. 
After solving, we will have an optimal control to apply corresponding to the correct instant in time $t_k+T_c$~\cite{koenemann_wholebody_2015}.

Finally, we cover the case where the refresh rate $\dt$ of the actuators' driver is (several times) faster than the node period, $\dt < \Dt$.
In this case, we can compute optimal controls every $\dt$ using the optimal control policy \eqref{eq:optimal_policy} provided with the last solution of the \gls*{fddp} solver.
For this, we first need to predict a state ${\bf x}(t_u)$ at the control instant $t_u$ using the dynamics model \eqref{eq:state_integration}, then compute a state error $\delta \bf x$ with respect to the estimated state at this same instant, $\delta {\bf x}=\hat {\bf x}(t_u) \ominus {\bf x}(t_u)$, and finally apply the control policy to find the controls.

After all these design considerations, we still have  a lot of freedom to specify the \glspl{ocp} at each time step.
In the following, we present three \gls{nmpc} strategies to build such \gls{ocp} problems from the mission specification.
The first two have been seen in the literature (although for tasks other than aerial manipulation), while the latter is a novelty of this paper.

\subsection{Weighted MPC}
\label{subsec:mpc_weighted}
\begin{figure}[t]
  \centering
  \includegraphics[trim=1.5em 0.0 0 0.0,clip,width=0.9\linewidth, keepaspectratio]{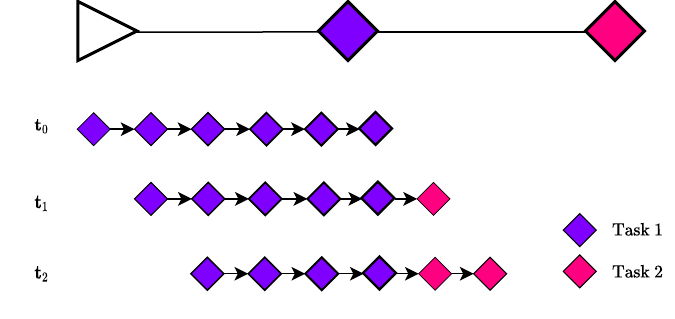}
  \caption{Weighted MPC controller. Top: mission specification. Bottom rows: successive \gls*{ocp} of the \gls*{wmpc} at three consecutive \gls*{nmpc} updates.  Nodes have the same cost function as the closest task to come. Their increasing weights are set according to \eqref{eq:weighted_mpc_weights}.}
  \label{fig:oc_weighted_mpc}
\end{figure}

This method is an adaptation from~\cite{kleff_highfrequency_2021}, which presents an \gls*{nmpc} controller to perform positioning tasks with a fully-actuated serial manipulator.
The structure of the \gls*{ocp} in the \gls*{wmpc} does not consider any node corresponding to a navigation phase.
It also avoids computing the reference trajectory completely.
Instead, each node adopts the cost function from the closest upcoming task (\figref{fig:oc_weighted_mpc}).
The costs' weights $w_{i,k}$ take exponentially increasing values as nodes approach the task,
\begin{align}
  \label{eq:weighted_mpc_weights}
  w_{i,k} &= w_i\,e^{\alpha(t_k-t_\text{task})}  
  \,,
\end{align}
where $w_i$ is the task weight given in the mission specification, $\alpha$ is a tuning parameter that controls the sharpness of the increasing cost, $t_k=t_0+k\Delta t$ is the time of the $k$-th \gls{ocp} node, and $t_\text{task}$ is the time when the next task starts.
Notice that some other functions might be valid too, \eg\ using a linear function or also assigning the weight according to a Gaussian envelope~\cite{neunert_fast_2016}.

The main advantage of \gls{wmpc} is that it is simple. 
It has however important limitations that can be understood from the effects of the tuning parameter $\alpha$, of the limited horizon, and of the absence of a proper warm start of the \gls{ocp} (the reference trajectory). 
With $\alpha$ large, the only nodes with significant weight are the ones really close to the task. 
This makes the other nodes behave as navigation nodes, thus allowing the \gls{ocp} to plan the optimal maneuvers to reach the task.
However, if the task is beyond the horizon, this automatically drops all the weights to zero: the optimal solution becomes a null torque on all motors with the \gls{am} falling to the ground.
This is handled by setting the horizon node's weight to the task's weight $w_i$, with the effect of bringing the task to one horizon's time distance, thus hurrying the controls to achieve it, at occasions making it unfeasible because of the impossibility to fit the appropriate maneuver. 
Notice that this strategy alters the optimality criterion set in the mission specification.
Another option is to decrease $\alpha$, thereby raising the weight of all the nodes within the horizon, now radically altering the mission's optimality criterion.
This has the adverse effect of eliminating the navigation nodes and therefore preventing non-trivial navigation maneuvers to emerge.\footnote{As a toy example, one can think of a pendulum with limited torque so that it has to perform a swing-up maneuver to drive the link to its upright equilibrium point. 
If the horizon is not long enough to accommodate the full swing-up maneuver, the controller will bring the link to a tilted position where the actuator torque equals the torque produced by the weight.}
Things get worse for under-actuated platforms because penalization costs on part of the state often prevent other parts to converge. 
For example, if the orientation during navigation is attracted to that of the task, then the platform tilts poorly and cannot accelerate to reach the destination.
This must be alleviated by what we call here \emph{\gls{uaw}}, that is, weighting through $\mathbf{W}_i$ in \eqref{eq:to_phase_cost_function} the different parts of the state differently (\eg, less weight on orientation than on position).
\gls{uaw}  requires careful tuning, which renders \gls{wmpc} less simple than its original version.
For all these reasons, this controller is only pertinent for fully-actuated platforms under low dynamics.

\subsection{Rail MPC}
\label{subsec:mpc_rail}
The \gls*{rmpc} is basically a tracking controller (like the one presented in~\cite{brunner_trajectory_2020}).
It tracks the reference trajectory $\mathbf{X}^* = \{\mathbf{x}_0^*, \mathbf{x}_1^*, \dots, \mathbf{x}_M^*\}$ that we get by solving a full-length \gls*{ocp} offline.
This trajectory acts as the rail where the controller runs, hence the name of this controller.
Notice that by solving this problem offline we already account for the high dynamics and the under-actuation of the system: the complex maneuvers will be discovered at the time of computing this reference trajectory.

\begin{figure}[t]
  \centering
  \includegraphics[trim=1.5em 110 0 0, clip, width=0.9\linewidth]{figures/oc_weighted_mpc.pdf}\\
  {\includegraphics[trim=1.5em 0.0 0 0.0,clip,width=0.9\linewidth]{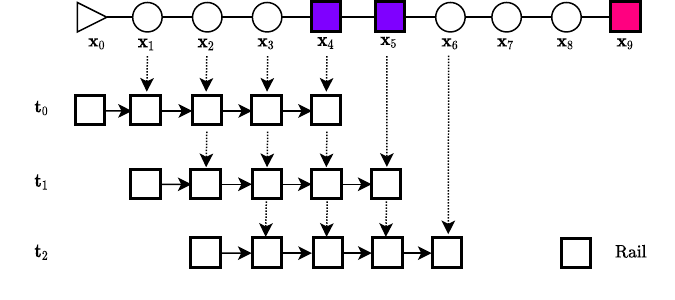}}
  \caption{Rail MPC controller. Top: mission specification. Mid: reference trajectory solved offline (circle: navigation node; square: task node). Bottom: successive \gls{ocp} problems. At every \gls*{nmpc} update the desired state of each residual is updated with the corresponding state from the reference trajectory. All nodes become task nodes devoted to follow the reference closely, regardless of any disturbances.}
  \label{fig:oc_rail_mpc}
\end{figure}

The structure of the \gls*{ocp} considers the same type of residual for every node: tasks residuals attracting the trajectory towards the reference rail, and regularization residuals for the control (see \figref{fig:oc_rail_mpc}), \ie,
\begin{equation}
  l_{rail, k}(\mathbf{x}_k, \mathbf{u}_k) = w_{\text{state}}\| \mathbf{r}_{\text{state}, k} \|^2_\mathbf{W_\text{state}} + w_{\text{ctrl}}\|\mathbf{r}_{\text{ctrl}, k}\|^2_\mathbf{W_\text{ctrl}} \,.
  \label{eq:rail_cost_function}
\end{equation}
See \secref{subsec:residuals} for more details on these residuals.

Using an \gls*{nmpc} controller as a tracking controller limits considerably its inherent re-planning capability, which is important in case of disturbances.
The structure of its cost function restricts the solution to be the one that passes through $\mathbf{X}^*$.
In the event of a disturbance, 
\gls*{rmpc} outputs a set of controls to return to the nominal trajectory. These controls are no longer optimal due to the disturbance.
Moreover, for under-actuated platforms and as it happened in \gls{wmpc}, we need to apply \gls{uaw} to allow good convergence after disturbances towards the reference trajectory. 

\subsection{Carrot MPC}
\label{subsec:mpc_carrot}

The \gls*{cmpc} tackles the drawbacks of \gls*{wmpc} and \gls*{rmpc}.
Departing also from a pre-computed reference trajectory $\bf X^*$ we want a strategy able to control a \gls*{am} in highly dynamic maneuvers and with the ability to reject disturbances in an optimal manner, \ie, by planning a new trajectory from the perturbed state, ignoring the reference trajectory.
This reference is only used to set the horizon node of the \gls{ocp}, which acts as the \emph{carrot} that the \gls{am} pursues.

\begin{figure}[t]
  \centering
  \includegraphics[trim=1.5em 110 0 0, clip, width=0.9\linewidth]{figures/oc_weighted_mpc.pdf}\\
  \includegraphics[trim=1.5em 0.0 0 0.0,clip,width=0.9\linewidth]{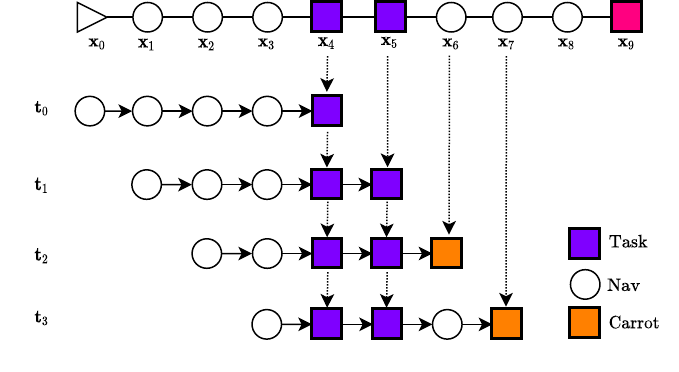}
  \caption{Carrot MPC controller. Top: mission specification. Mid: reference trajectory solved offline (circle: navigation node; square: task node). Bottom: successive \gls{ocp} problems. The carrot task node (orange square) is used to encode the optimal cost-to-go value function related to the original problem. Navigation nodes in the \gls*{ocp} provide the  slackness needed to replan a new optimal trajectory in the event of a disturbance.}
  \label{fig:oc_carrot_mpc}
\end{figure}

The idea behind the \gls*{cmpc} draws inspiration from the Bellman's principle of optimality. 
This states that any optimal solution to a problem can be divided into several sub-problems whose solutions coincide with the original one.
We apply it here by considering that beyond the horizon the optimal trajectory is given by the reference $\bf X^*$, while to reach the horizon we truly recompute a new optimal trajectory at each time (\figref{fig:oc_carrot_mpc}).
The horizon node taken from the reference acts as the carrot.
Through a task residual with sharp quadratic cost, this carrot node encodes the \emph{cost-to-go} value function \eqref{eq:optimal_ctg} from the horizon to the end of the mission.
Notice that, unlike \gls{wmpc} and \gls{rmpc}, all except the task and carrot nodes are true navigation nodes.
The part of trajectory with such nodes is \emph{slack}, \ie, arbitrary states or controls for these nodes have a small penalization in the  cost function.
This gives the controller the ability to recompute, online, a new optimal trajectory towards the carrot or task states, whichever comes first, eventually discovering new maneuvers, with an optimality criterion that matches the one originally specified for the mission, and without any need for \gls{uaw} or any other tuning.


\urlstyle{rm}

\section{Validation} \label{sec:validation}

The versatility of optimal control allows us to use it within a wide range of applications and situations involving \glspl*{am} (see \secref{subsec:contribution}).
We present in \secref{subsec:validation_to} a number of examples involving different scenarios of trajectory optimization for full-body torque-controlled \glspl*{am}.
See \tabref{tab:experiments_summary} for a summary of the trajectories.
Then in \secref{subsec:validation_mpc} we close the control loop to test the performance of the different \gls*{nmpc} controllers described in \secref{sec:mpc}.
%
%

All \glspl*{ocp} in these experiments have been built with an open source \textsc{C++} library delivered with this paper, called \textsc{EagleMPC}\footnote{\url{https://github.com/PepMS/eagle-mpc}}.
As a dependency to build and solve the \gls*{ocp}, we use \textsc{Crocoddyl}~\cite{mastalli_crocoddyl_2020}, which in turn uses \textsc{Pinocchio}~\cite{carpentier_pinocchio_2019} to compute the fast articulated body dynamics algorithms based on~\cite{featherstone_rigidbody_book} such as the \gls*{aba}.
\textsc{EagleMPC} includes the implementation of the \gls*{nmpc} controllers explained in \secref{sec:mpc} as well as the implementation of the SquashBox FDDP solver presented in~\cite{marti_squash-box_2020}.
Besides, we also provide specific tools to ease the assembly of optimization problems for \glspl*{am}, such as a YAML file parser to specify problems following the philosophy explained in \secref{sec:cost_functions}.

The simulations in \secref{subsec:validation_mpc} use the \textsc{Gazebo}\footnote{\url{http://gazebosim.org/}} simulator, and a modified version of \textsc{RotorS}\footnote{\url{https://github.com/PepMS/rotors_simulator}} \cite{furrer_rotors_2016}, which has been specialized to aerial manipulation.
The Robot Operating System (\textsc{ROS})\footnote{\url{http://ros.org/}} packages needed to run \gls*{nmpc} controllers along with some other useful tools are also provided\footnote{\url{https://github.com/PepMS/eagle-mpc-ros}}.
The rest of software dependencies along with further instructions on how to reproduce the experiments are collected in a specific repository\footnote{\url{https://github.com/PepMS/fbtl_nmpc_experiments}}.
All these experiments have run single-threaded in a laptop with an Intel® Core™ i7-8750H CPU @ 2.20GHz $\times$ 12 with 32GB of RAM.

\subsection{Trajectory optimization}
\label{subsec:validation_to}

In this section, we describe how we build  \glspl*{ocp} for different missions with the costs considered for each phase.
To showcase the abilities of the presented approach, we include challenging missions with the following characteristics (usually not considered in the state-of-the-art of aerial manipulation):
\begin{itemize}
  \item \textbf{Aggressiveness}: trajectories with motions ranging from quasi-static (slow) to agile (fast) maneuvers.
  \item \textbf{Type of aerial manipulators:} we combine 3 different platforms, including a non-planar hexacopter~\cite{rajappa_tilthex_2015}, with 3 serial manipulators of different DoFs (see \tabref{tab:experiments_summary}).
  \item \textbf{Contacts:} we present some simulations with contacts between the aerial robot and the environment.
  \item \textbf{Model changes:} we provoke some sudden mass changes during the maneuvers.
\end{itemize}

\begin{figure*} [t]
  \centering\begin{tabular}{lclc}
    \rowname{a} & \raisebox{-.5\height}{\href{https://youtu.be/K176KlK7Gbk}{\includegraphics[trim=0 100 0 170, clip, width=0.4\linewidth]{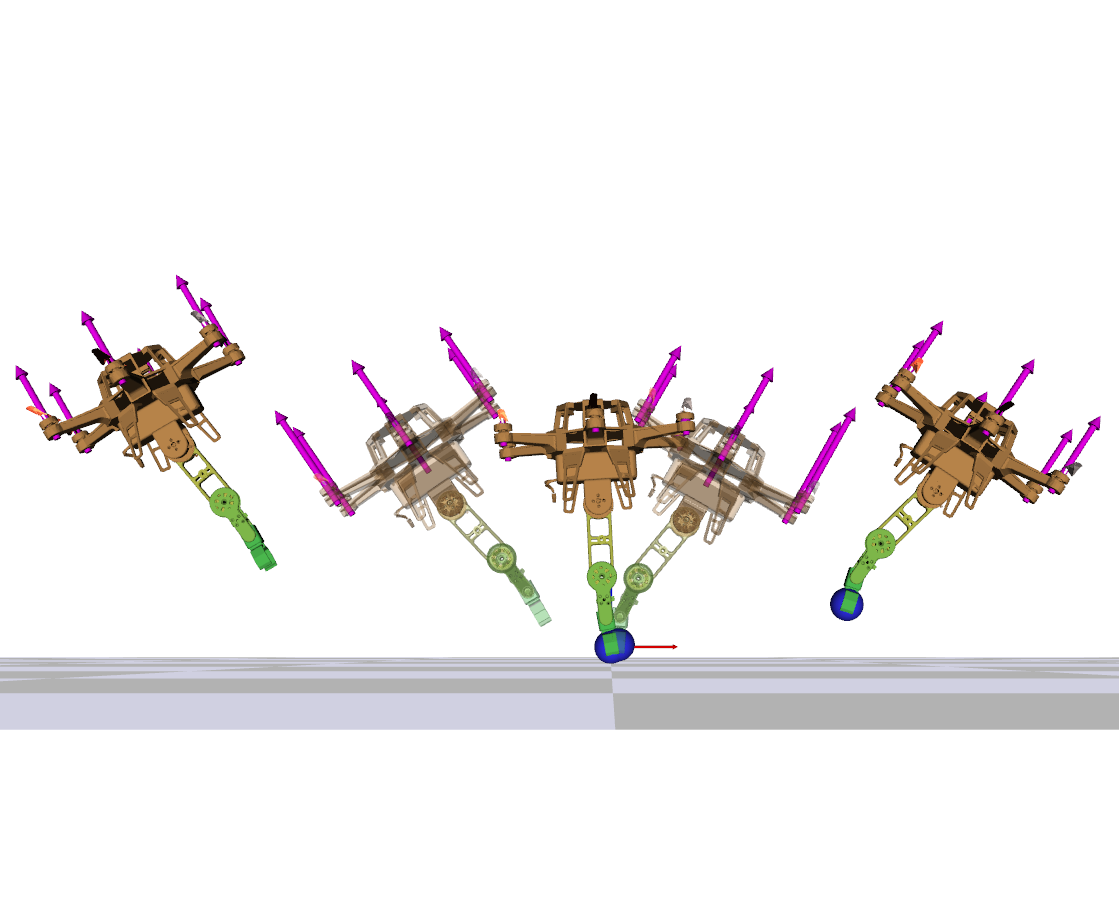}}}~~~~~~~~ &
    \rowname{b} & \raisebox{-.5\height}{\href{https://youtu.be/ASaQKXnYHZc}{\includegraphics[trim=0 100 0 170, clip, width=0.43\linewidth]{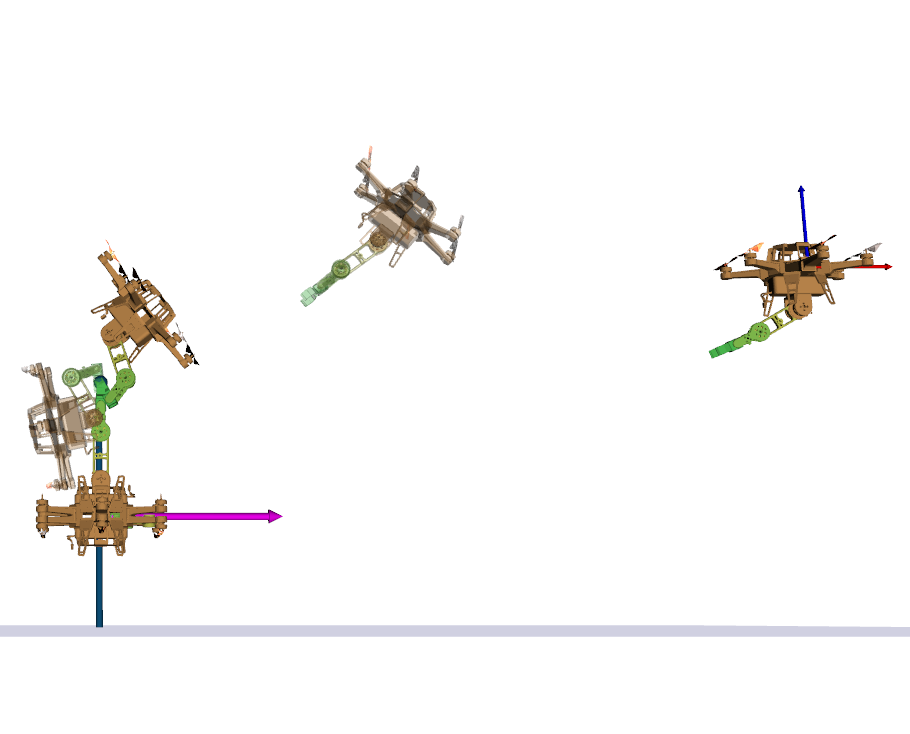}}}           \\
    \midrule
    \rowname{c} & \raisebox{-.5\height}{\href{https://youtu.be/gdxz1hZhaE4}{\includegraphics[trim=0 100 0 20, clip, width=0.4\linewidth]{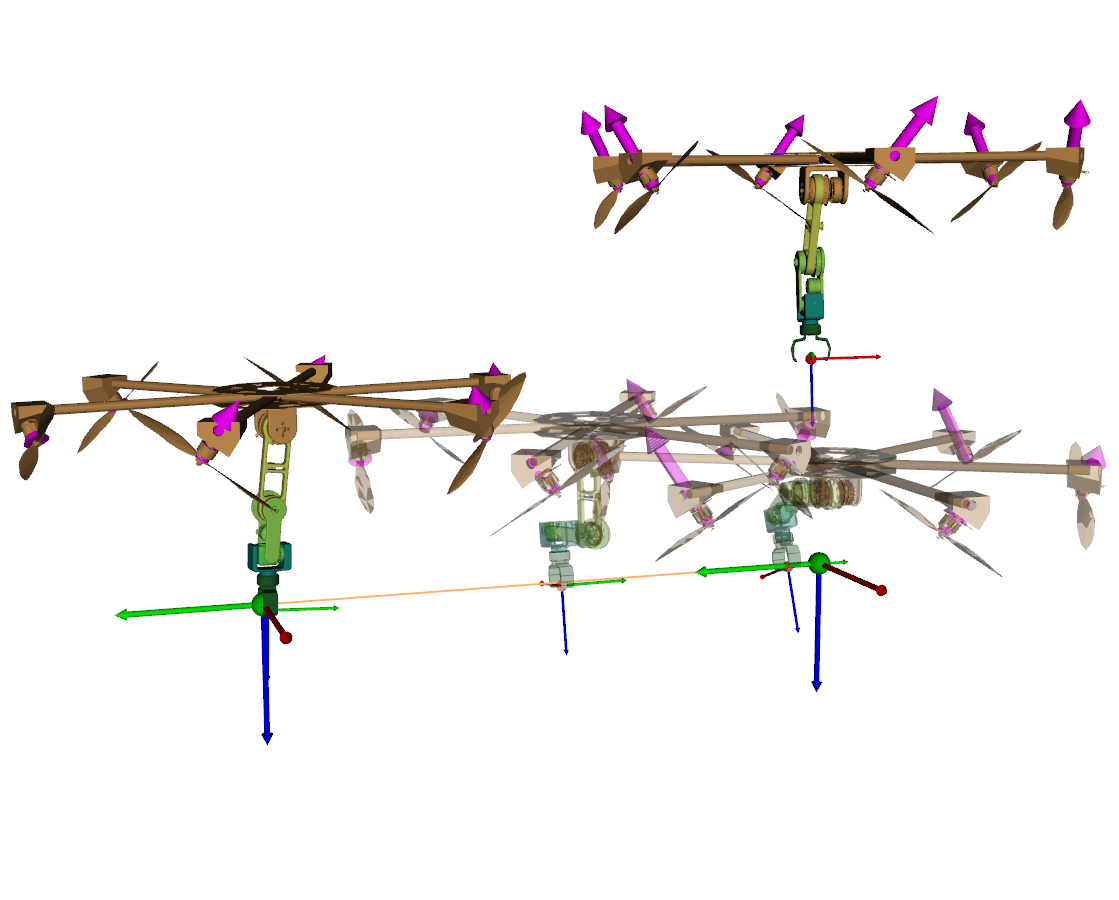}}}~~~~~~~~   &
    \rowname{d} & \raisebox{-.5\height}{\href{https://youtu.be/cPBolWN3C-Q}{\includegraphics[trim=0 0 0 20, clip, width=0.4\linewidth]{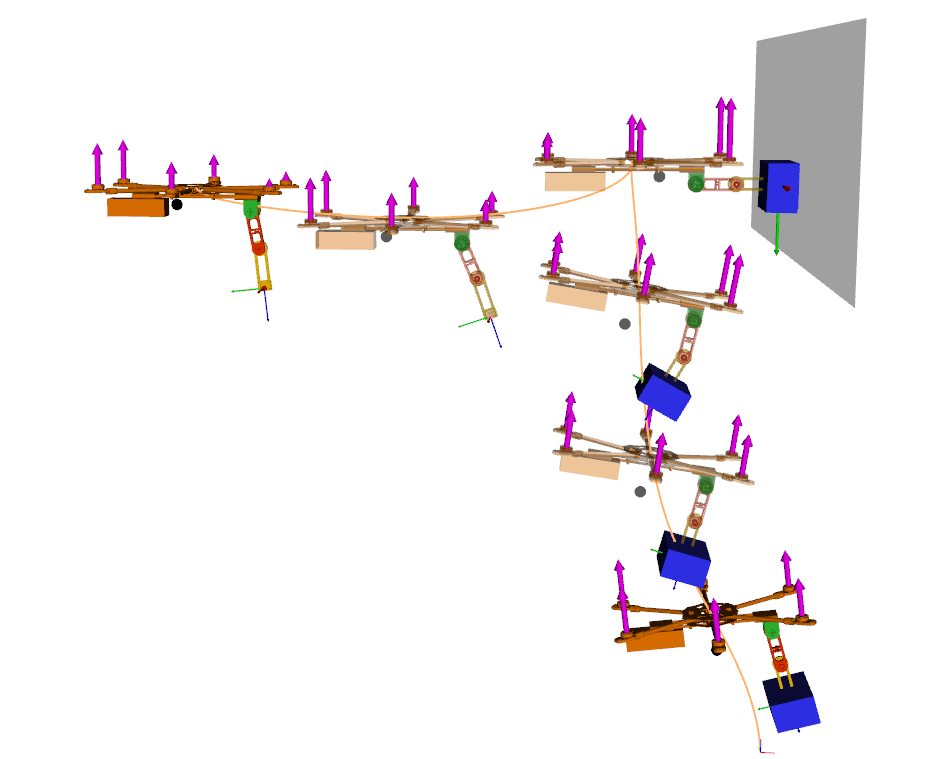}}}           \\
    \midrule
  \end{tabular}
  \caption{Sequence of each mission.
    (a) \emph{Eagle's Catch} with Hexacopter370 \& 3\gls*{dof} arm. (b) Monkey-bar with Hexacopter370 \& 3\gls*{dof} arm. (c) Push \& slide with Tilthex \& 5\gls*{dof} arm. (d) Box deployment with Hexacopter680 \& 2\gls*{dof} arm. Click on images to see videos.}
  \label{fig:uam_to_maneuvers}
\end{figure*}

To compute these trajectories we cold-start the solver, \ie, the initial guess is a trajectory where all states and controls are zero.
The platform and mission characteristics are detailed in \tabref{tab:experiments_summary}.
Mission phases and costs are in \tabref{tab:ocp_definition}.
The configuration of the \gls{ocp} and its main performance indices come in \tabref{tab:ocp_solver_performance}.
The four missions are displayed in \figref{fig:uam_to_maneuvers} --- we strongly encourage the reader to look at the provided videos to appreciate the dynamics and full-body control.
They are described and evaluated hereafter.

\begin{table}[t]
  \caption{Mission configurations and characteristics}
  \label{tab:experiments_summary}
  \begin{center}
    \begin{tabular}{@{}l c cccc@{}}
      \toprule
                                          &  & \rotatebox{90}{Eagle's catch} & \rotatebox{90}{Monkey bar} & \rotatebox{90}{Push \& slide} & \rotatebox{90}{Box depolyment} \\
      \midrule
      \textsc{Platforms}                                                                                                                                                   \\
      Hexacopter 370 (planar)             &  & \cmark                        & \cmark                                                                                      \\
      Hexacopter 670 (planar)             &  &                               &                            &                               & \cmark                         \\
      Tilthex (non-planar)                &  &                               &                            & \cmark                        &                                \\
      \midrule
      \textsc{Manipulators}                                                                                                                                                \\
      Serial link manipulator-2\gls*{dof} &  &                               &                            &                               & \cmark                         \\
      Serial link manipulator-3\gls*{dof} &  & \cmark                        & \cmark                                                                                      \\
      Serial link manipulator-5\gls*{dof} &  &                               &                            & \cmark                                                         \\
      \midrule
      \textsc{Mission characteristics}                                                                                                                                     \\
      Aggressive                          &  & \cmark                        & \cmark                                                                                      \\
      Contacts                            &  & \cmark                        & \cmark                     & \cmark                        &                                \\
      Model changes                       &  &                               &                            &                               & \cmark                         \\
      \bottomrule
    \end{tabular}
  \end{center}
\end{table}

\begin{table}[t]
  \caption{\glspl*{ocp}' definitions. Costs specified for \textsc{bl}: base link; \textsc{ee}: end effector; \textsc{l1}: link 1. Phases \textsc{N}: navigation; \textsc{T}: task; \textsc{C}: contact.}
  \label{tab:ocp_definition}
  \begin{center}
    \begin{tabular}{@{} l c cccccc @{}}
      \toprule                   &                                                 & \multicolumn{6}{c}{\textsc{Residual types}}                                                                                                                                                                                                           \\
      \cmidrule{3-8}                                                                                                                                                                                                                                                                                                                       \\
                                 & \rotatebox{90}{$T_\text{phase}$ [\si{\second}]} & \rotatebox{90}{$\mathbf{x}, \mathbf{u}$ reg. \&  $\mathbf{x}$ bounds} & \rotatebox{90}{Frame pose} & \rotatebox{90}{Frame position} & \rotatebox{90}{Frame orientation} & \rotatebox{90}{Frame velocity} & \rotatebox{90}{Friction cone: \textsc{ee}} \\
      \midrule
      \textsc{Eagle's catch}                                                                                                                                                                                                                                                                                                               \\
      Phase 1: approach (N)      & $1.4$                                           & \cmark                                                                                                                                                                                                                                                \\
      Phase 2: pre-catch (T, C)  & -                                               & \cmark                                                                &                            & \textsc{ee}                    &                                   & \textsc{ee}                                                                 \\
      Phase 3: catch (T, C)      & $0.1$                                           & \cmark                                                                &                            & \textsc{ee}                    &                                   & \textsc{ee}                    & \cmark                                     \\
      Phase 4: fly away (N)      & $1.6$                                           & \cmark                                                                                                                                                                                                                                                \\
      Phase 5: hover (T)         & -                                               & \cmark                                                                &                            & \textsc{bl}                    &                                   & \textsc{bl}                    &                                            \\

      \midrule
      \textsc{Monkey bar}                                                                                                                                                                                                                                                                                                                  \\
      Phase 1: balancing (N, C)  & $1.4$                                           & \cmark                                                                &                            & \textsc{ee}                                                                                                                                      \\
      Phase 2: flying (N)        & $0.5$                                           & \cmark                                                                                                                                                                                                                                                \\
      Phase 3: hover (T)         & -                                               & \cmark                                                                & \textsc{bl}                                                                                                                                                                   \\
      \midrule
      \textsc{Push \& Slide}                                                                                                                                                                                                                                                                                                               \\
      Phase 1: approach (N)      & $1.5$                                           & \cmark                                                                                                                                                                                                                                                \\
      Phase 2: pre P \& S (T)    & -                                               & \cmark                                                                & \textsc{ee}                & \textsc{bl}                    & \textsc{bl}                       & \textsc{ee}, \textsc{bl}                                                    \\
      Phase 3: P \& S (T, C)     & $2.5$                                           & \cmark                                                                & \textsc{ee}                & \textsc{bl}                    & \textsc{bl}                       & \textsc{ee}                                                                 \\
      Phase 4: P \& S end (T, C) & $-$                                             & \cmark                                                                & \textsc{ee}                & \textsc{bl}                    & \textsc{bl}                       & \textsc{ee}, \textsc{bl}                                                    \\
      \midrule
      \textsc{Box Deployment}                                                                                                                                                                                                                                                                                                              \\
      Phase 1: approach (N)      & $2.0$                                           & \cmark                                                                                                                                                                                                                                                \\
      Phase 2: anochoring (T)    & $1.0$                                           & \cmark                                                                & \textsc{ee}                &                                & \textsc{l1}                       & \textsc{ee}                                                                 \\
      Phase 3: fly-away (N)      & $2.0$                                           & \cmark                                                                                                                                                                                                                                                \\
      Phase 4: hover (T)         & -                                               & \cmark                                                                & \textsc{bl}                &                                &                                   & \textsc{bl}                                                                 \\
      \bottomrule
    \end{tabular}
  \end{center}
\end{table}

\begin{table}[t]
  \caption{Solver performance}
  \label{tab:ocp_solver_performance}
  \begin{center}
    \begin{tabular}{@{}l c cccc@{}}
      \toprule
                                                              & \rotatebox{90}{Eagle's catch} & \rotatebox{90}{Monkey bar} & \rotatebox{90}{Push \& slide} & \rotatebox{90}{Box depolyment} \\
      \midrule
      Duration $T_M~[\si{\second}]$                           & $3.1$                         & $1.9$                      & $4$                           & $5$                            \\
      Node period $\Dt~[\si{\milli\second}]$                  & $20$                          & $5$                        & $20$                          & $20$                           \\
      \# Nodes  $N$                                           & $156$                         & $381$                      & $202$                         & $254$                          \\
      Integrator                                              & SI-E$^1$                      & RK4$^2$                    & RK4                           & SI-E                           \\
      \# Iterations                                           & $22$                          & $181$                      & $80$                          & $29$                           \\
      Solving time $[\si{\second}]$                           & $0.21$                        & $11.9$                     & $4.4$                         & $0.25$                         \\
      \quad per iteration $[\si{\milli\second}]$              & $9.1$                         & $65.7$                     & $55.0$                        & $8.6$                          \\
      \qquad per node $[\si{\micro\second}]$                  & $58$                          & $173$                      & $272$                         & $34$                           \\
      \qquad\quad per \# controls cubed $[\si{\nano\second}]$ & $80$                          & $237$                      & $204$                         & $66$                           \\
      \bottomrule
      $^1$SI-E: Semi-implicit Euler                                                                                                                                                         \\
      $^2$RK4: $4$-th order Runge-Kutta
    \end{tabular}
  \end{center}
\end{table}

\subsubsection{Eagle's catch}

This mission emulates an eagle catching an animal on the ground.
A detail of the catching maneuver is shown in \figref{fig:uam_to_maneuvers}a.
The first part of the trajectory is meant to drive the end-effector to the catching point (phases 1 and 2).
There, it transitions to the catching phase (phase 3), which contains a contact between the end-effector and the ground.
This contact is modeled as a non-sliding 3D contact point to constrain the translation, allowing rotations around it.
To ensure the validity of the contact modeling and avoid slippage, we have added a cost that penalizes forces out of the friction cone.
Finally, the robot flies away towards a hovering point further on (phases 4 and 5).

\begin{figure}[t]
  \centering
  \includegraphics[trim=7.0 0.0 5.0 5.0,clip,width=\linewidth]{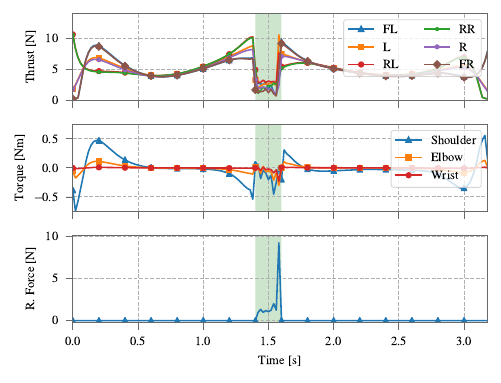}
  \caption{\emph{Eagle's Catch} mission. Top: platform motor thrust; Center: torque of the arm motors; Bottom: $L_2$ norm of the reaction force at the end-effector. Shaded area indicates the catching phase.}
  \label{fig:eagle_catch_plot}
\end{figure}

The resulting control trajectory shows a remarkable maneuver discovered by the \gls{ocp} during the catching phase (see Figs.~\ref{fig:cover}, \ref{fig:uam_to_maneuvers}a and \ref{fig:eagle_catch_plot}).
Notice how the thrusts of the motors decay considerably, while the arm holds a small fraction of the \gls{am}'s weight in a quasi-singular configuration (arm straight, near-zero torques).
The \gls{am} is acting partially as a passive inverted pendulum, partially as in a ballistic parabolic path (observable in \figref{fig:cover}), consuming almost no energy.
While this behavior might seem unnatural, it is however optimal given the specification.
Two different approaches to a more realistic eagle's catch would be to impose a minimum flight velocity during the catch phase (a must in fixed-wing platforms), or to specify a contactless mission.
We explore this last possibility in the \gls{nmpc} experiments in \secref{subsec:validation_mpc}.

\subsubsection{Monkey bar}

Inspired by~\cite{delamare_toward_2018}, the aim of this experiment is to validate contacts as a useful source of motion for \glspl*{am}.
The experiment considers a \gls*{am} that is initially hanging upside down from a bar which it grasps with the end-effector.
The objective is to bring the platform to a waypoint that is placed in front of the robot and above the ground (see \figref{fig:uam_to_maneuvers}b).
We simulate an overloaded \gls{am} by limiting the maximum thrust of its 6 propellers to $4$\si{\newton}.
This is insufficient to hold its total weight of $25$\si{\newton}, and thus to be able to reach the waypoint the robot has to jump from the bar with its arm.
The contact with the bar is modeled as a contact point that constrains the three linear \glspl*{dof}.

\begin{figure}[t]
  \centering
  \includegraphics[trim=5.0 0.0 5.0 5.0,clip,width=\linewidth]{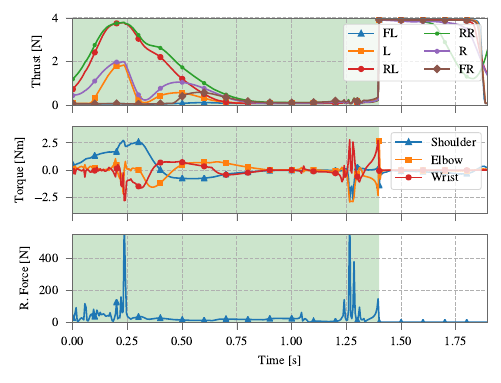}
  \caption{Monkey bar mission. Top: platform motor thrust; Center: torque of the arm motors; Bottom: $L_2$ norm of the reaction force at the end-effector. Shaded area indicates the contact phase, which has two distinguished parts separated by a resting stage: climbing to the upright position, and jumping at $t=1.4\si{\second}$.}
  \label{fig:monkey_bar_plot}
\end{figure}

%
The first phase includes the robot climb on top of the bar and the jump.
\figref{fig:monkey_bar_plot} shows how, from $0$\si{\second} to approximately  $0.75$\si{\second}, all motors are used to produce the climb to an upright position.
From there, the robot gently leans forward to a final push to eventually jump from the bar (from $1$\si{\second} to approximately  $1.4$\si{\second}).
Notice how at the jump instant the arm is producing a maximum push  against the bar of almost $200\si{\newton}$.
At this point the robot enters the flying phase, during which the actuators saturate at full throttle to hold the weight of the \gls{am}.
As the \gls{am} approaches the waypoint they orient the platform so it can reach it with the desired pose.
This pose corresponds to the only task in this trajectory (phase 3).

Interestingly, as the contact phase is significantly long, it may occur that the end-effector separates from the contact point due to a numerical drift when integrating the constraint $\ddot{\phi}(\mathbf{q}) = \mathbf{0}$.
To ensure the numerical stability of the contact we have chosen a small integration step of $\Delta t = 5$\si{\milli\second} along with a \gls*{rk} integrator of 4-th order.
This is a simple and sufficient approach here.
One can use other techniques  to improve the numerical stability for such constrained trajectories such as the addition of Baumgarte's gains~\cite{blajer_methods_2011}.

\subsubsection{Tilthex push \& slide}
This mission is meant to validate two key concepts.
First, we want to show that the proposed formulation can deal with complex \gls*{am} designs.
We use for this a \gls*{am} based on the fully-actuated \emph{Tilthex} platform~\cite{rajappa_tilthex_2015} with a non-planar 5\glspl*{dof} manipulator (see \figref{fig:uam_to_maneuvers}c).
The second idea is to check that an \gls*{ocp} approach can effectively solve interacting tasks that have been tackled so far with different approaches.
For this, we  choose a typical interaction problem, the so called \emph{push \& slide}, by which we push on a surface with the end-effector while sliding it along a straight line path.

\begin{figure}[t]
  \centering
  \includegraphics[trim=0.0em 0.0 5.0 5.0,clip,width=\linewidth]{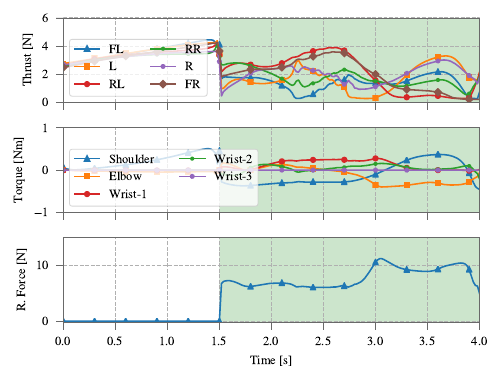}
  \caption{Push \& Slide mission. Top: platform motor thrust; Center: torque of the arm motors; Bottom: $L_2$ norm of the reaction force at the end-effector. Shaded area indicates the actual push \& slide phase.}
  \label{fig:push_slide_plot}
\end{figure}

The mission starts by bringing the end-effector at the beginning of the push \& slide path (phases 1 and 2).
Phase 2 finishes when the robot is at disposal of starting the pushing phase.
That is, the end-effector has a specific pose with a null velocity and the platform is parallel to the ground at specific height (see the costs of phase 2 in \tabref{tab:ocp_definition}).
The height requirement for the platform is to make sure that the arm is away from singularities.
The push \& slide begins at phase 3 with the end-effector in contact with the surface (indicated by the green shaded area in \figref{fig:push_slide_plot}).
This contact is modeled as a rolling contact along the forward (y) axis, constraining the linear movements in the lateral (x) and vertical (z) axes.
To ease the convergence of the solver, we have added a cost penalizing the end-effector for deviating from the slide path.

\subsubsection{Box deployment}

This experiment validates the use of the \gls*{ocp} framework for situations where the model varies.
We simulate the deployment of a $1$\si{\kilo\gram} self-anchoring package onto a wall (see \figref{fig:uam_to_maneuvers}d).
Here, in order to ensure the proper placement of the box, we hold it at the same anchoring place for one second (mimicking a wall contact as in the case of a wall-grasping action).
The deployment involves a switch between two different models, \ie, with and without the box, implying important mass changes during the trajectory (\ie, different parameters of the dynamic model).

\begin{figure}[t]
  \centering
  \includegraphics[trim=0.0 0.0 5.0 5.0,clip,width=\linewidth]{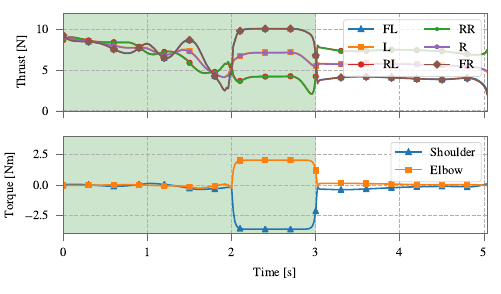}
  \caption{Box deployment. Top: platform motor thrust; Bottom: torque of the arm motors. Shaded area indicates the phase carrying the load.}
  \label{fig:box_deployment_plot}
\end{figure}

We use a planar hexacopter with a planar 2\gls*{dof} arm.
A first navigation phase leads the end-effector and the box towards the anchoring pose.
To avoid the collision between the propellers and the wall, we must ensure a fully extended configuration when the end-effector reaches the box deployment phase (phase 2).
For this purpose, we have added a cost to enforce the orientation of link 1.
During this phase, (from second two to second three) we can see in \figref{fig:box_deployment_plot} how the front motors have to compensate for the displaced mass.
The model switching occurs at the end of this phase 2 and the platform motors have to account for the sudden change of the center of mass of the robot (non-shaded area in \figref{fig:box_deployment_plot}).

\subsection{Model Predictive Controllers}
\label{subsec:validation_mpc}

The aim of this section is to test the \gls*{nmpc} controllers presented in \secref{sec:mpc} and verify experimentally the respective behaviors that we have anticipated.

In this section we do not deal with hybrid problems, which involve model switches due to contacts or changes of mass.
Solving these problems within an \gls*{nmpc} framework requires additional reasoning and decisions that are out of the scope of this paper.
For example, in the case of contacts we have to decide whether we feed the \gls{ocp} with a predefined contact sequence or, differently, we endow the \gls*{nmpc} with the ability to decide the placements for the contacts (\cite{neunert_wholebody_2018}).
We could also allow for unplanned contacts to occur (with the result of impacts that would render the derivatives discontinuous) and implement a contact detector to, online, modify the dynamics model at the precise contact switching instants.
These and other important algorithmic additions are not obvious and would deviate the focus of the present work.
We believe that considering single-model problems suffices to validate the \gls*{nmpc} strategies that we have described.

All controllers share the same basic configuration.
Their \glspl*{ocp} have $N=30$ nodes separated $\Delta t = 30\si{\milli\second}$, which results in a horizon of $T_H=870\si{\milli\second}$, and are limited to $n=4$ iterations.
Robot state estimates arrive every $\dt=2.5\si{\milli\second}$ (\ie, $400 \si{\hertz}$), triggering the \gls*{ocp} update and solving.
We have tuned each controller  to perform at its best.
In particular, we use \gls{uaw} in \gls*{rmpc} and \gls{wmpc}, where the weights for the platform position and the arm joints' angles are set significantly higher than those of orientation and velocity. 

\subsubsection{Underactuation}

We set two experiments, \emph{4-Displacement} and \emph{Eagle's Catch},   with different complexities to test how each controller copes with the under-actuation of the platform, including slow and aggressive maneuvers.

\begin{figure}[t]
  \centering
  \href{https://youtu.be/BmHJksA9Sog}{\includegraphics[width=\linewidth]{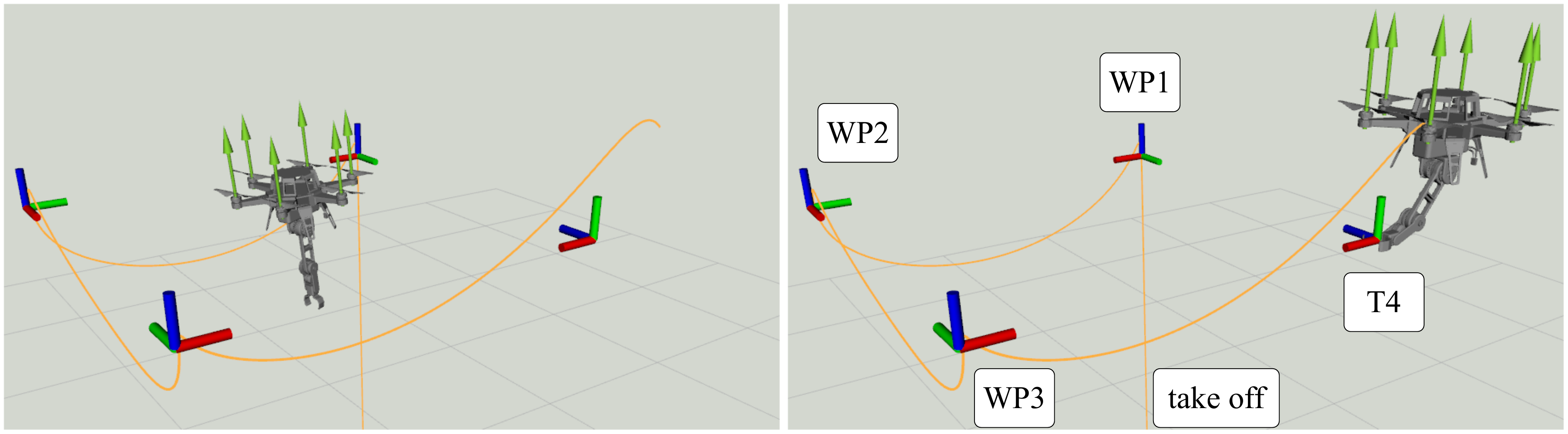}}
  \href{https://youtu.be/BmHJksA9Sog}{\includegraphics[trim=10.0 0.0 6.0 0.0,clip,width=\linewidth]{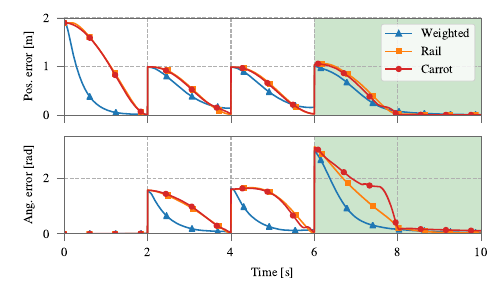}
  }
  \caption{The \emph{4-Displacement} mission. Intermediate and final snapshots, and pose error of the incoming task for the three \gls*{nmpc} controllers. Pose waypoints WP\{1,2,3\} for the platform at times $t=\{2,4,6\}\si{\second}$. Shaded area indicates a pose task T4 for the end effector at $t=8\si{\second}$. Click on images to see video.}
  \label{fig:mpc_4displacement_task_error}
\end{figure}

The \emph{4-Displacement} mission (\figref{fig:mpc_4displacement_task_error}) requires slow movements.
It has four tasks to be accomplished at seconds 2, 4, 6 and 8, which are separated by navigation phases.
The first three tasks WP1--3 specify a desired pose for the platform.
The last task T4 specifies a desired pose for the end-effector and a desired hovering orientation for the platform.
\figref{fig:mpc_4displacement_task_error} shows the error norm of the pose residuals, with the three controllers managing to reduce the error for each task (notice that the error from $6\si{\second}$ to $10\si{\second}$ is related to the end-effector's desired pose).
While \gls*{rmpc} and \gls*{cmpc} have similar levels of task accomplishments, the \gls*{wmpc} shows larger position errors.
This is because \gls*{wmpc} adds the residuals of the task to every node in the horizon (see \figref{fig:oc_weighted_mpc}):
the controls hurry to get to the task, which is good but far from optimal, but then struggle to truly converge due to insufficient \gls*{uaw} tuning.
Since there are no significant disturbances, \gls{rmpc} and \gls{cmpc} execute the reference trajectory, which is optimal, with similar results. \gls{cmpc} can however react at the last moment to better reposition the end effector, a maneuver visible before second 8 in \figref{fig:mpc_4displacement_task_error}.

\begin{figure}[t]
  \centering
  \href{https://youtu.be/dX7IFYdaDPA}{\includegraphics[width=\linewidth]{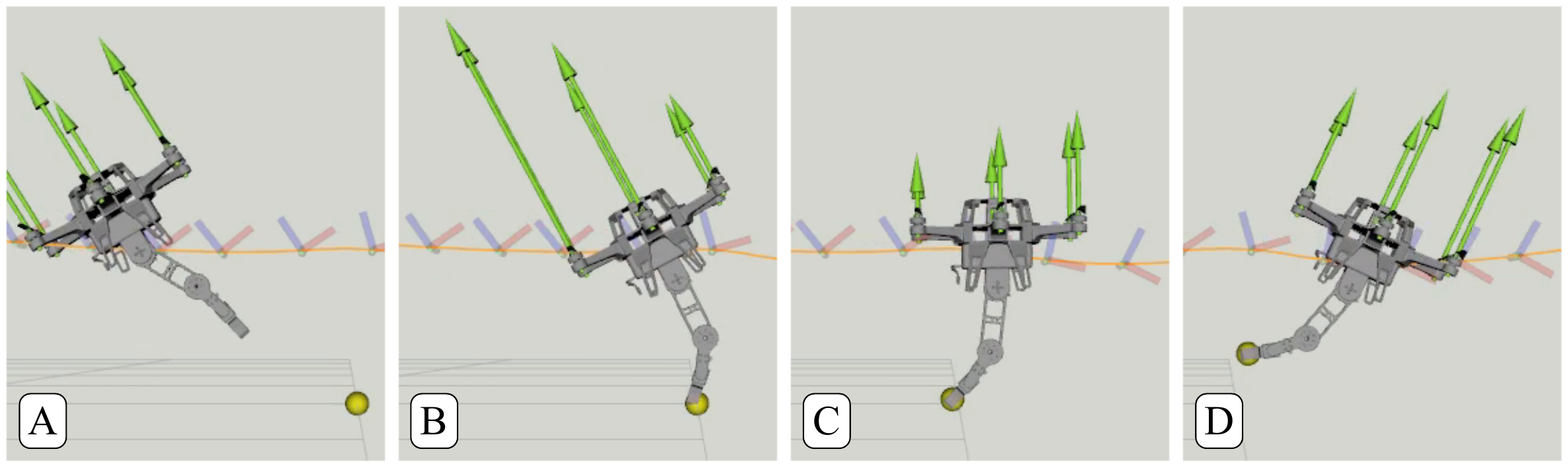}}
  \href{https://youtu.be/dX7IFYdaDPA}{\includegraphics[trim=0.0 0.0 6.0 0.0,clip,width=\linewidth]{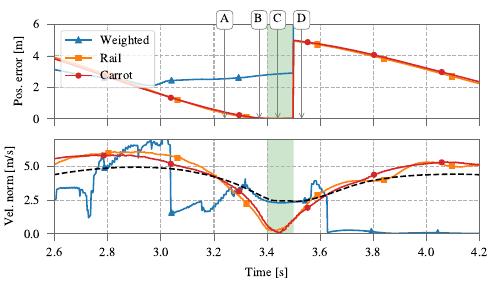}}
  \caption{The \emph{Eagle's Catch} mission. Top: snapshots around the catch instant labeled A, B, C, D. Center: end effector position error during the approaching and catching phase. Bottom: Linear velocity norms of end effector (colors) and platform (dashed black, only \gls{cmpc} is shown). Shaded areas indicate the catching phase. Notice the anticipation and swing of the arm to be able to reach the object with the end effector at zero velocity while the platform flies at $2.5\si{\meter}/\si{\second}$. Notice also that \gls{cmpc} is smoother. Click on images to see video.}
  \label{fig:mpc_eagle_catch_task_error}
\end{figure}

The performance differences between the three controllers are magnified when the trajectory becomes more aggressive.
Hence, here we evaluate them with an \emph{Eagle's Catch} mission 
where this time
we are not considering the contact with the ground.
\figref{fig:mpc_eagle_catch_task_error} shows some snapshots around the catch task and the norm of the residuals from each controller. 
In this experiment, \gls*{wmpc} fails to reach the grasping point due to its lack of prediction capability (see \secref{subsec:mpc_weighted}), while \gls*{rmpc} and \gls*{cmpc} behave satisfactorily.

\subsubsection{Disturbance rejection}

\begin{figure}[t]
  \centering
  \includegraphics[trim=0.0 25 5.0 5.0,clip,width=\linewidth]{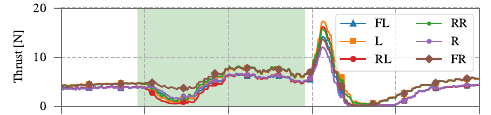}\\
  \includegraphics[trim=0.0 0.0 5.0 0.0,clip,width=\linewidth]{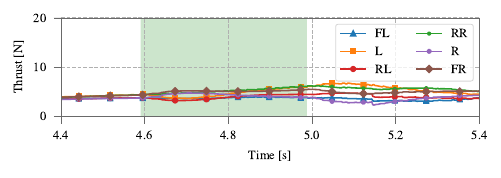}
  \caption{Thrust control signals for the \emph{4-Displacement} mission during the  \emph{well before task} disturbance. Top: \gls{rmpc}; Bottom: \gls{cmpc}. Shaded area indicates the disturbance is active. The motor reactions to the disturbance are almost imperceptible in \gls{cmpc}.}
  \label{fig:mpc_disturbances_trajectories}
\end{figure}

As seen in the previous experiments, with a proper \gls{uaw} tuning of the weights in the \gls*{rmpc}, both \gls*{rmpc} and \gls*{cmpc} can accomplish tasks with a similar precision to that of the offline computed trajectory.
This is an expected behavior in nominal conditions with only small perturbations.
However, differences between these two emerge when larger disturbances enter the scene.
To evidence these, we have run three simulation case studies where three disturbances of $10\si{\newton}$ in the $[1,1,0]$ direction  are applied to the robot during $0.4\si{\second}$ while performing the \emph{4-Displacement} mission.
These disturbances come just before the task WP2, during the task WP2, and well before the next task WP3.
\begin{figure}[t]
  \centering
  \href{https://youtu.be/nKmM0mOUYXA}{\includegraphics[trim=5.0 4.0 5.0 18.0,clip,width=\linewidth]{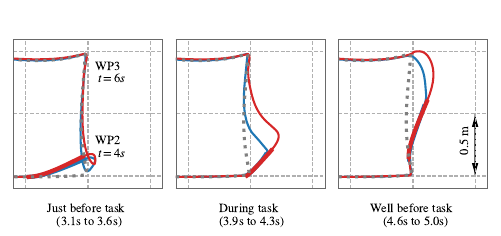}}
  \caption{XY trajectories for the \emph{4-Displacement} experiment with disturbances starting at different times. Dotted gray: reference trajectory; Blue: \gls{rmpc}; Red: \gls{cmpc}. Thicker portion indicates the disturbance is active. While the \gls*{rmpc} tries to return to the reference trajectory, the \gls*{cmpc} re-plans a new optimal trajectory towards the next task. See \figref{fig:mpc_disturbances_trajectories} for the corresponding control signals. Click on images to see video.}
  \label{fig:mpc_disturbances_xy}
\end{figure}
The resulting platform trajectories are shown in~\figref{fig:mpc_disturbances_xy}.
We see how  \gls*{rmpc} always tries to track the reference trajectory while  \gls*{cmpc} re-plans a new optimal trajectory to fulfill  the  tasks.
This results in much more aggressive commands for the \gls*{rmpc} (shown in \figref{fig:mpc_disturbances_trajectories}), which work hard for regaining the reference trajectory even though this is not a requirement of the mission itself.
That is, \gls*{cmpc} consumes much less energy than \gls*{rmpc} to fulfill the same task.

For the sake of completeness and to confirm the behaviors anticipated in \figref{fig:mpc_disturbances_xy}, we have performed a Monte Carlo simulation consisting of 50 experiments where a random step disturbance is applied for each controller and each starting moment.
The duration $\Dt$ and the magnitude $F$ of the disturbance have been determined by randomly sampling the respective Gaussian distributions, $\Delta t\,[\si{\second}] \sim \mathcal{N}(0.5\,, 0.25^2 )$ and $F\,[\si{\newton}] \sim \mathcal{N}(8\,, 2^2)$.
The starting time $t_0$ has been sampled from uniform distributions with ranges $[3.0, 3.2]$, $[3.7, 3.9]$ and $[4.4, 4.6]$ for the `before', `during' and `well before' cases, respectively.
The force is applied along the $[1,1,0]$ direction in the inertial coordinates.
\begin{figure}[t]
  \centering
  \includegraphics[trim=0.5em 0.0 0 0.0,clip,width=\linewidth]{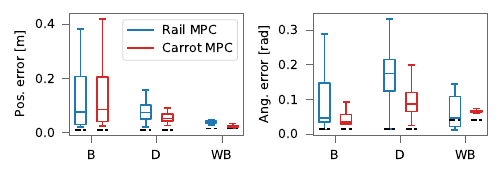}
  \caption{Task error distributions for \gls*{rmpc} and \gls*{cmpc}. B: before task WP2. D: during WP2. WB: well before WP3. Dashed black lines indicate the error of the reference trajectory.}
  \label{fig:mpc_box_disturbance}
\end{figure}
We show in~\figref{fig:mpc_box_disturbance} the resulting distributions of the norm of the position and orientation errors of the concerned tasks.
We can see that \gls*{cmpc} is generally more accurate (less error) and more predictable (less variance) than \gls*{rmpc}.
Notice that the orientation errors are larger in \gls*{rmpc} than in \gls*{cmpc}; this is due to \gls*{uaw} in \gls*{rmpc}, which weights orientations less than position.
Notice also the very predictable orientation error of \gls*{cmpc} in the WB case: 
while this error is in great part due to an error of the reference trajectory, this indicates that the disturbances far enough from tasks do not alter the optimal way to fulfill the task.

Finally, it is worth mentioning that none of the \gls*{nmpc} controllers described in this work is capable of rejecting persistent disturbances (notice that all simulated disturbances are of temporal action).
This is because the disturbances are not part of the model and the controller cannot account for the discrepancy between the model and the reality if they are persistent.
We believe that an external force observer would certainly improve this, however we consider it out of scope by now and leave its implementation as future work.

\subsubsection{Solver performance}
Any \gls*{nmpc} controller relies on an \gls{ocp} solver that must be able to do one or more iterations in a very short time.
In \tabref{tab:solver_performance}, we show some statistics related to the solving time per \gls*{nmpc} step, both in the \emph{4-Displacement} and the \emph{Eagle's Catch} missions.
Notice that all the maximum values fit in the \gls{nmpc} step of $\Dt=30\si{\milli\second}$, thus enabling true \gls{nmpc} control.
However, highly dynamic platforms can run more smoothly and stably if the solver outputs a solution at higher rates, \eg\ every time it receives an estimation of the state and before receiving the next one.
In our $400\si{\hertz}$ system this means  solving in less than $2.5\si{\milli\second}$.
As shown in \tabref{tab:solver_performance}, this is not always possible, as some of the means and all the maxima are above $2.5\si{\milli\second}$.
In all the experiments of this paper the  \gls*{nmpc} applies the control once the \gls*{ocp} has finished iterating and waits for the next state estimate to solve the next \gls*{nmpc} step.
This suboptimal strategy has been proven sufficiently stable and accurate.
We have described in \secref{subsec:architecture} a few possible improvements.
Please refer to the conclusions below for further discussion.

\begingroup
\setlength{\tabcolsep}{4.5pt}
\begin{table}[t]
  \caption{Solver performance. N: nominal, D: with disturbance}
  \label{tab:solver_performance}
  \begin{center}
    \begin{tabular}{@{} l ccc c ccc@{}}
      \toprule
                                & \multicolumn{3}{c}{\textsc{Solving time} [\si{\milli\second}]} &         & \multicolumn{3}{c}{\# \textsc{Iterations}}                            \\
      \cmidrule{2-4}  \cmidrule{6-8}
                                & Mean                                                           & St. dev & Max.                                       &  & Mean & St. dev & Max. \\
      \midrule
      \textsc{Rail MPC}                                                                                                                                                            \\
      \emph{4-Displacement} (N) & 2.58                                                           & 1.02    & 18.4                                       &  & 1.52 & 0.51    & 4    \\
      \emph{4-Displacement} (D) & 2.92                                                           & 1.13    & 15.4                                       &  & 1.49 & 0.51    & 4    \\
      \emph{Eagle's Catch} (N)  & 2.02                                                           & 0.78    & 10.5                                       &  & 1.24 & 0.43    & 3    \\
      \textsc{Carrot MPC}                                                                                                                                                          \\
      \emph{4-Displacement} (N) & 2.40                                                           & 1.02    & 19.3                                       &  & 1.23 & 0.44    & 4    \\
      \emph{4-Displacement} (D) & 2.77                                                           & 1.09    & 15.5                                       &  & 1.23 & 0.44    & 4    \\
      \emph{Eagle's Catch} (N)  & 2.28                                                           & 1.02    & 17.8                                       &  & 1.20 & 0.45    & 4    \\
      \bottomrule
    \end{tabular}
  \end{center}
\end{table}
\endgroup



\section{Conclusions}
\label{sec:conclusions}

We have proposed a full methodology to control \glspl*{am} based on full-body, torque-level, model predictive control.
This methodology has been imported from the humanoids and legged robots community and adapted to the \gls*{am}.
Such adaptation comprises some original contributions, particularly, in the methods to close the control loop through \gls*{nmpc}, which seem to be the most poorly explored by the original communities.
We have explained the different parts in a way that can be regarded as a tutorial, for it is complete but avoids deviating too much from the main line of work constituting the current state of the art.
These deviations range from techniques to add hard constraints to the solver, considering impacts in the treatment of contacts, the joint optimization of the timings in the mission plans, the online identification of the model and/or any persistent disturbance, or others.
Some of these techniques can be explored in the provided references, while others might require new research.

The methods presented here have already proved their pertinence in the  fields of humanoid and legged robotics with great success.
In the area of \gls*{am}, in this paper we have boarded the mathematical and software aspects, and have provided evidence of their maturity and fitness using realistic simulations that cover a wide range of platforms and challenging missions.

These techniques have yet to be demonstrated in real UAMs, and a few comments regarding our confidence in this proposal are necessary.
First, we tested the algorithms in their basic versions and succeeded in generating accurate and stable motor commands for all experiments.
We presented in \secref{subsec:architecture} possible improvements that should allow us to go beyond the performances displayed here.
Second, we have run the algorithms single threaded in a standard CPU.
Today, low weight CPUs suited for UAM such as the Intel\textsuperscript{\textregistered}\,NUC\,11\footnote{\url{https://www.intel.com/content/www/us/en/products/details/nuc.html}} are already more powerful than our test bench CPU, and the FDDP algorithm accepts multithreading.
These two aspects make us confident in the fact that processing power should not be an issue when moving to real \glspl*{am}, and we believe we can extend the prediction horizon beyond one second.
Third and of special importance is tackling the electromechanics of the real implementation. 
This actually constitutes our immediate future work.
Regarding the flying platform, accessing the motor drivers to send propeller speed commands can be done through open-source platforms such as the Pixhawk-4\footnote{\url{https://docs.px4.io/master/en/flight_controller/pixhawk4.html}} controller and PIX4\footnote{\url{https://px4.io/}} libraries.
The manipulator arms require significantly more attention, especially regarding light and high-torque motors, miniature power electronics and light arm structures.
Fortunately, this way has been paved already by the legged robots community, for example through activities such as the Open Dynamic Robot Initiative\footnote{\url{https://open-dynamic-robot-initiative.github.io/}}~\cite{grimminger2020open} which proposes open-source torque-controlled 3D-printed links with torques up to 2.7Nm and weights below the 150g, based on which different arm configurations can be built.
An important conclusion of the present study is that these arm specifications are sufficient to accomplish all the missions that we have simulated, and therefore that we can make real torque-controlled UAMs work with hardware (CPUs, motors, electronics and structures) available today.


\section*{Acknowledgments}
The authors would like to thank Nicolas Mansard at LAAS-CNRS in Toulouse and Carlos Mastalli at the University of Edinburgh for their support to get us started in the field of optimal control.
Also, thanks to Arash Kalantari (NASA-JPL), for the collaboration in the project \emph{Deployable self-adhering anchoring platform and its deployment system for future Mars Helicopters}, exploiting some of the concepts presented in this paper.
Also, special thanks to Ricard Bordalba (IRI, CSIC-UPC) for his valuable feedback.



\bibliography{./files/root}

\begin{thebibliography}{10}
\providecommand{\url}[1]{#1}
\csname url@samestyle\endcsname
\providecommand{\newblock}{\relax}
\providecommand{\bibinfo}[2]{#2}
\providecommand{\BIBentrySTDinterwordspacing}{\spaceskip=0pt\relax}
\providecommand{\BIBentryALTinterwordstretchfactor}{4}
\providecommand{\BIBentryALTinterwordspacing}{\spaceskip=\fontdimen2\font plus
\BIBentryALTinterwordstretchfactor\fontdimen3\font minus
  \fontdimen4\font\relax}
\providecommand{\BIBforeignlanguage}[2]{{%
\expandafter\ifx\csname l@#1\endcsname\relax
\typeout{** WARNING: IEEEtran.bst: No hyphenation pattern has been}%
\typeout{** loaded for the language `#1'. Using the pattern for}%
\typeout{** the default language instead.}%
\else
\language=\csname l@#1\endcsname
\fi
#2}}
\providecommand{\BIBdecl}{\relax}
\BIBdecl

\bibitem{carpentier_leggedrobots_2021}
J.~Carpentier and P.~Wieber, ``\href{https://hal.inria.fr/hal-03193886}{Recent
  Progress in Legged Robots Locomotion Control},'' \emph{{HAL - Current
  Robotics Reports}}, 2021.

\bibitem{betts_practicalmethods_optimalcontrol_book}
J.~Betts,
  \emph{\href{https://epubs.siam.org/doi/abs/10.1137/1.9780898718577}{Practical
  Methods for Optimal Control and Estimation Using Nonlinear Programming}},
  2nd~ed.\hskip 1em plus 0.5em minus 0.4em\relax Cambridge University Press,
  2009.

\bibitem{ruggiero_am_literature_review_2018}
F.~Ruggiero, V.~Lippiello, and A.~Ollero, ``Aerial manipulation: A literature
  review,'' \emph{IEEE Robot. Automat. Lett.}, 2018.

\bibitem{khamseh_am_survey_2018}
K.~H. B., F.~Janabi-Sharifi, and A.~Abdessameud,
  ``\href{http://www.sciencedirect.com/science/article/pii/S0921889017305535}{Aerial
  manipulation: A literature survey},'' \emph{Robotics and Autonomous Systems},
  2018.

\bibitem{ruggiero_multilayer_2015}
F.~Ruggiero, M.~A. Trujillo, R.~Cano, H.~Ascorbe, A.~Viguria, C.~Perez,
  V.~Lipiello, A.~Ollero, and B.~Siciliano,
  ``\href{https://ieeexplore.ieee.org/document/7139760?arnumber=7139760}{A
  multilayer control for multirotor UAVs equipped with a servo},'' in
  \emph{IEEE Int. Conf. Rob. Autom. (ICRA)}, 2015.

\bibitem{rossi_trajgeneration_2017}
R.~Rossi, A.~Santamaria-Navarro, J.~Andrade-Cetto, and P.~Rocco,
  ``\href{https://ieeexplore.ieee.org/document/7762762}{Trajectory Generation
  for Unmanned Aerial Manipulators Through Quadratic Programming},'' \emph{IEEE
  Robot. Automat. Lett.}, 2017.

\bibitem{lipiello_visualservoing_2016}
V.~Lippiello, J.~Cacace, A.~Santamaria-Navarro, J.~Andrade-Cetto, M.~A.
  Trujillo, Y.~R. Rodríguez~Esteves, and A.~Viguria,
  ``\href{https://ieeexplore.ieee.org/abstract/document/7361979}{Hybrid Visual
  Servoing With Hierarchical Task Composition for Aerial Manipulation},''
  \emph{IEEE Robot. Automat. Lett.}, 2016.

\bibitem{rajappa_tilthex_2015}
S.~Rajappa, M.~Ryll, H.~H. Bülthoff, and A.~Franchi, ``Modeling, control and
  design optimization for a fully-actuated hexarotor aerial vehicle with tilted
  propellers,'' in \emph{IEEE Int. Conf. Rob. Autom. (ICRA)}, 2015.

\bibitem{tognon_control-aware_2018}
M.~Tognon, E.~Cataldi, H.~A.~T. Chavez, G.~Antonelli, J.~Cortés, and
  A.~Franchi,
  ``\href{https://ieeexplore.ieee.org/document/8283709}{Control-aware Motion
  Planning for Task-Constrained Aerial Manipulation},'' \emph{IEEE Robot.
  Automat. Lett.}, 2018.

\bibitem{kim_stabilizing_2018}
M.~J. Kim, K.~Kondak, and C.~Ott,
  ``\href{https://ieeexplore.ieee.org/document/8283724}{A Stabilizing
  Controller for Regulation of UAV With Manipulator},'' \emph{IEEE Robot.
  Automat. Lett.}, 2018.

\bibitem{murray_differentialflatness_1995}
R.~M. Murray, M.~Rathinam, and W.~Sluis,
  ``\href{http://www.cds.caltech.edu/~murray/preprints/mrs95-imece.pdf}{Differential
  Flatness of Mechanical Control Systems: A Catalog of Prototype Systems},'' in
  \emph{ASME Int. Congress and Exposition}, 1995.

\bibitem{fliess_differentialflatness_1999}
M.~Fliess, J.~Levine, P.~Martin, and P.~Rouchon,
  ``\href{https://ieeexplore.ieee.org/document/763209}{A Lie-Backlund approach
  to equivalence and flatness of nonlinear systems},'' \emph{IEEE Tran. on
  Automatic Control}, 1999.

\bibitem{mellinger_minimum_2011}
D.~Mellinger and V.~Kumar,
  ``\href{https://ieeexplore.ieee.org/document/5980409?arnumber=5980409}{Minimum
  snap trajectory generation and control for quadrotors},'' in \emph{IEEE Int.
  Conf. Rob. Autom. (ICRA)}, 2011.

\bibitem{sreenath_cablesuspended_2013}
K.~Sreenath, N.~Michael, and V.~Kumar,
  ``\href{https://ieeexplore.ieee.org/document/6631275}{Trajectory generation
  and control of a quadrotor with a cable-suspended load - A
  differentially-flat hybrid system},'' in \emph{IEEE Int. Conf. Rob. Autom.
  (ICRA)}, 2013.

\bibitem{yuksel_differential_2016}
B.~Yüksel, G.~Buondonno, and A.~Franchi,
  ``\href{https://ieeexplore.ieee.org/document/7759109?arnumber=7759109}{Differential
  flatness and control of protocentric aerial manipulators with any number of
  arms and mixed rigid-/elastic-joints},'' in \emph{IEEE/RSJ Int. Conf. Intell.
  Rob. Sys. (IROS)}, 2016.

\bibitem{tognon_dynamic_2017}
M.~Tognon, B.~Yüksel, G.~Buondonno, and A.~Franchi,
  ``\href{https://ieeexplore.ieee.org/abstract/document/7989753}{Dynamic
  decentralized control for protocentric aerial manipulators},'' in \emph{IEEE
  Int. Conf. Rob. Autom. (ICRA)}, 2017.

\bibitem{ryll_6dinteraction_2019}
M.~Ryll, G.~Muscio, F.~Pierri, E.~Cataldi, G.~Antonelli, F.~Caccavale,
  D.~Bicego, and A.~Franchi,
  ``\href{https://journals.sagepub.com/doi/abs/10.1177/0278364919856694}{6D
  Interaction Control with Aerial Robots: The Flying End-Effector Paradigm},''
  \emph{Int. J. of Rob. Research}, 2019.

\bibitem{hamaza_adaptivecompliance_2018}
S.~Hamaza, I.~Georgilas, and T.~Richardson,
  ``\href{https://ieeexplore.ieee.org/document/8452382?arnumber=8452382}{An
  Adaptive-Compliance Manipulator for Contact-Based Aerial Applications},'' in
  \emph{IEEE/ASME Int. Conf. Adv. Intell. Mechatronics (AIM)}, 2018.

\bibitem{tognon_trulyredundant_2019}
M.~Tognon, H.~A.~T. Chávez, E.~Gasparin, Q.~Sablé, D.~Bicego, A.~Mallet,
  M.~Lany, G.~Santi, B.~Revaz, J.~Cortés, and A.~Franchi,
  ``\href{https://ieeexplore.ieee.org/document/8629273}{A Truly-Redundant
  Aerial Manipulator System With Application to Push-and-Slide Inspection in
  Industrial Plants},'' \emph{IEEE Robot. Automat. Lett.}, 2019.

\bibitem{delamare_toward_2018}
Q.~Delamare, P.~R. Giordano, and A.~Franchi, ``Toward aerial physical
  locomotion: The contact-fly-contact problem,'' \emph{IEEE Robot. Automat.
  Lett.}, 2018.

\bibitem{lunni_nonlinear_2017}
D.~Lunni, A.~Santamaria-Navarro, R.~Rossi, P.~Rocco, L.~Bascetta, and
  J.~Andrade-Cetto,
  ``\href{https://ieeexplore.ieee.org/document/7991347}{Nonlinear model
  predictive control for aerial manipulation},'' in \emph{IEEE Int. Conf.
  Unman. Aircr. (ICUAS)}, 2017.

\bibitem{garimella_towards_2015}
G.~Garimella and M.~Kobilarov,
  ``\href{https://ieeexplore.ieee.org/document/7139850}{Towards
  model-predictive control for aerial pick-and-place},'' in \emph{IEEE Int.
  Conf. Rob. Autom. (ICRA)}, 2015.

\bibitem{geisert_trajectory_2016}
M.~Geisert and N.~Mansard,
  ``\href{https://ieeexplore.ieee.org/document/7487460}{Trajectory generation
  for quadrotor based systems using numerical optimal control},'' in \emph{IEEE
  Int. Conf. Rob. Autom. (ICRA)}, 2016.

\bibitem{tzoumanikas_aerial_2020}
D.~Tzoumanikas, F.~Graule, Q.~Yan, D.~Shah, M.~Popovic, and S.~Leutenegger,
  ``\href{http://www.roboticsproceedings.org/rss16/p046.pdf}{Aerial
  manipulation using hybrid force and position NMPC applied to aerial
  writing},'' in \emph{Rob.: Sci. Sys. (RSS)}, 2020.

\bibitem{kamel_fastnonlinear_2015}
M.~Kamel, K.~Alexis, M.~Achtelik, and R.~Siegwart, ``Fast nonlinear model
  predictive control for multicopter attitude tracking on so(3),'' in
  \emph{IEEE Conf. on Contr. Applications (CCA)}, 2015.

\bibitem{neunert_fast_2016}
M.~Neunert, C.~d. Crousaz, F.~Furrer, M.~Kamel, F.~Farshidian, R.~Siegwart, and
  J.~Buchli, ``\href{https://ieeexplore.ieee.org/document/7487274}{Fast
  nonlinear model predictive control for unified trajectory optimization and
  tracking},'' in \emph{IEEE Int. Conf. Rob. Autom. (ICRA)}, 2016.

\bibitem{brunner_trajectory_2020}
M.~Brunner, K.~Bodie, M.~Kamel, M.~Pantic, W.~Zhang, J.~Nieto, and R.~Siegwart,
  ``Trajectory tracking nonlinear model predictive control for an overactuated
  mav,'' in \emph{IEEE Int. Conf. Rob. Autom. (ICRA)}, 2020.

\bibitem{brescianini_omni_2016}
D.~Brescianini and R.~D'Andrea, ``Design, modeling and control of an
  omni-directional aerial vehicle,'' in \emph{IEEE Int. Conf. Rob. Autom.
  (ICRA)}, 2016.

\bibitem{nocedal_numerical_book}
J.~Nocedal and S.~J. Wright,
  \emph{\href{https://www.springer.com/gp/book/9780387303031}{Numerical
  Optimization}}, 2nd~ed.\hskip 1em plus 0.5em minus 0.4em\relax Springer,
  2006.

\bibitem{wachter_mp_2006}
A.~W{\"a}chter and L.~T. Biegler,
  ``\href{https://link.springer.com/article/10.1007/s10107-004-0559-y#citeas}{On
  the implementation of an interior-point filter line-search algorithm for
  large-scale nonlinear programming},'' \emph{Mathematical Programming}, 2006.

\bibitem{byrd_knitro_2006}
R.~H. Byrd, J.~Nocedal, and R.~A. Waltz,
  ``\href{http://citeseerx.ist.psu.edu/viewdoc/summary?doi=10.1.1.126.425}{KNITRO:
  An integrated package for nonlinear optimization},'' in \emph{Large Scale
  Nonlinear Optimization, 35–59, 2006}, 2006.

\bibitem{mayne_secondorder_1966}
D.~Mayne, ``\href{https://doi.org/10.1080/00207176608921369}{A Second-order
  Gradient Method for Determining Optimal Trajectories of Non-linear
  Discrete-time Systems},'' \emph{Int. J. of Contr.}, 1966.

\bibitem{li_ilqr_2004}
W.~Li. and E.~Todorov.,
  ``\href{https://www.scitepress.org/ProceedingsDetails.aspx?ID=DnUUVSAoAUI=&t=1}{Iterative
  Linear Quadratic Design For Nonlinear Biological Movement Systems},'' in
  \emph{Int. Conf. on Inf. in Contr., Autom. and Rob. (ICINCO)}, 2004.

\bibitem{sideris_slq_2005}
A.~Sideris and J.~E. Bobrow,
  ``\href{https://ieeexplore.ieee.org/document/1470308}{An efficient sequential
  linear quadratic algorithm for solving nonlinear optimal control problems},''
  in \emph{IEEE American Control Conf.}, 2005.

\bibitem{neunert_wholebody_2018}
M.~Neunert, M.~Stäuble, M.~Giftthaler, C.~D. Bellicoso, J.~Carius, C.~Gehring,
  M.~Hutter, and J.~Buchli,
  ``\href{https://ieeexplore-ieee-org.recursos.biblioteca.upc.edu/document/8276298}{Whole-Body
  Nonlinear Model Predictive Control Through Contacts for Quadrupeds},''
  \emph{IEEE Robot. Automat. Lett.}, 2018.

\bibitem{koenemann_wholebody_2015}
J.~Koenemann, A.~Del~Prete, Y.~Tassa, T.~E., O.~Stasse, M.~Bennewitz, and
  N.~Mansard, ``\href{10.1109/IROS.2015.7353843}{Whole-body model-predictive
  control applied to the HRP-2 humanoid},'' in \emph{IEEE/RSJ Int. Conf.
  Intell. Rob. Sys. (IROS)}, 2015.

\bibitem{giftthaler_family_2018}
M.~Giftthaler, M.~Neunert, M.~Stäuble, J.~Buchli, and M.~Diehl,
  ``\href{https://ieeexplore.ieee.org/document/8593840}{A Family of Iterative
  Gauss-Newton Shooting Methods for Nonlinear Optimal Control},'' in
  \emph{IEEE/RSJ Int. Conf. Intell. Rob. Sys. (IROS)}, 2018.

\bibitem{mastalli_crocoddyl_2020}
C.~Mastalli, R.~Budhiraja, W.~Merkt, G.~Saurel, B.~Hammoud, M.~Naveau,
  J.~Carpentier, L.~Righetti, S.~Vijayakumar, and N.~Mansard,
  ``\href{https://ieeexplore.ieee.org/document/9196673}{Crocoddyl: An Efficient
  and Versatile Framework for Multi-Contact Optimal Control},'' in \emph{IEEE
  Int. Conf. Rob. Autom. (ICRA)}, 2020.

\bibitem{howell_altro_2019}
T.~A. Howell, B.~E. Jackson, and Z.~Manchester,
  ``\href{https://ieeexplore.ieee.org/document/8967788}{ALTRO: A Fast Solver
  for Constrained Trajectory Optimization},'' in \emph{IEEE/RSJ Int. Conf.
  Intell. Rob. Sys. (IROS)}, 2019.

\bibitem{mastalli_boxfddp_2020}
C.~Mastalli, W.~Merkt, J.~Marti{-}Saumell, J.~Sol{\`{a}}, N.~Mansard, and
  S.~Vijayakumar, ``\href{https://arxiv.org/abs/2010.00411}{A Direct-Indirect
  Hybridization Approach to Control-Limited {DDP}},'' \emph{CoRR}, 2020.

\bibitem{kazdadi_eqddp_2021}
S.~E. Kazdadi, J.~Carpentier, and J.~Ponce,
  ``\href{https://hal.inria.fr/hal-03184203}{Equality Constrained Differential
  Dynamic Programming},'' in \emph{IEEE Int. Conf. Rob. Autom. (ICRA)}, 2021.

\bibitem{marti_squash-box_2020}
J.~Mart\'i-Saumell, J.~Sol\`a, C.~Mastalli, and A.~Santamaria-Navarro,
  ``\href{https://cmastalli.github.io/publications/squashddp20iros.html}{Squash-Box
  Feasibility Driven Differential Dynamic Programming},'' in \emph{IEEE/RSJ
  Int. Conf. Intell. Rob. Sys. (IROS)}, 2020.

\bibitem{sola_micro_2018}
J.~Sol\`a, J.~Deray, and D.~Atchuthan,
  ``\href{http://arxiv.org/abs/1812.01537}{A micro {Lie} theory for state
  estimation in robotics},'' \emph{arXiv:1812.01537 [cs]}, 2018.

\bibitem{rawlings_modelpredictivecontrol_book}
J.~B. Rawlings, D.~Q. Mayne, and M.~Diehl,
  \emph{\href{https://sites.engineering.ucsb.edu/~jbraw/mpc/}{Model predictive
  control: theory, computation, and design}}.\hskip 1em plus 0.5em minus
  0.4em\relax Nob Hill Publishing, 2017.

\bibitem{featherstone_rigidbody_book}
R.~Featherstone,
  \emph{\href{http://www.springer.com/gp/book/9780387743141}{Rigid body
  dynamics algorithms}}.\hskip 1em plus 0.5em minus 0.4em\relax Springer, 2008.

\bibitem{budhiraja_ddp_multiphase_2018}
R.~Budhiraja, J.~Carpentier, C.~Mastalli, and N.~Mansard,
  ``\href{https://ieeexplore.ieee.org/document/8624925}{Differential Dynamic
  Programming for Multi-Phase Rigid Contact Dynamics},'' in \emph{IEEE Int.
  Conf. Hum. Rob. (ICHR)}, 2018.

\bibitem{diehl_fastdirect_2006}
M.~Diehl, H.~Bock, H.~Diedam, and P.-B. Wieber, \emph{Fast Motions in
  Biomechanics and Robotics: Optimization and Feedback Control}.\hskip 1em plus
  0.5em minus 0.4em\relax Berlin, Heidelberg: Springer Berlin Heidelberg, 2006,
  ch. \href{https://doi.org/10.1007/978-3-540-36119-0_4}{Fast Direct Multiple
  Shooting Algorithms for Optimal Robot Control}, pp. 65--93.

\bibitem{kleff_highfrequency_2021}
S.~Kleff, A.~Meduri, R.~Budhiraja, N.~Mansard, and L.~Righetti,
  ``High-frequency nonlinear model predictive control of a manipulator,'' in
  \emph{IEEE Int. Conf. Rob. Autom. (ICRA)}, 2021.

\bibitem{carpentier_pinocchio_2019}
J.~Carpentier, G.~Saurel, G.~Buondonno, J.~Mirabel, F.~Lamiraux, O.~Stasse, and
  N.~Mansard, ``\href{https://ieeexplore.ieee.org/document/8700380/}{The
  Pinocchio C++ library : A fast and flexible implementation of rigid body
  dynamics algorithms and their analytical derivatives},'' in \emph{IEEE Int.
  Sym. System Integration (SII)}, 2019.

\bibitem{furrer_rotors_2016}
F.~Furrer, M.~Burri, M.~Achtelik, and R.~Siegwart, \emph{Robot Operating System
  (ROS): The Complete Reference (Volume 1)}.\hskip 1em plus 0.5em minus
  0.4em\relax Cham: Springer International Publishing, 2016, ch.
  \href{http://dx.doi.org/10.1007/978-3-319-26054-9_23}{RotorS---A Modular
  Gazebo MAV Simulator Framework}, pp. 595--625.

\bibitem{blajer_methods_2011}
W.~Blajer,
  ``\href{https://www.sciencedirect.com/science/article/pii/S0045782511000089}{Methods
  for constraint violation suppression in the numerical simulation of
  constrained multibody systems – A comparative study},'' \emph{Computer
  Methods in Applied Mechanics and Engineering}, 2011.

\bibitem{grimminger2020open}
F.~{Grimminger}, A.~{Meduri}, M.~{Khadiv}, J.~{Viereck}, M.~{W\"uthrich},
  M.~{Naveau}, V.~{Berenz}, S.~{Heim}, F.~{Widmaier}, T.~{Flayols}, J.~{Fiene},
  A.~{Badri-Spr\"owitz}, and L.~{Righetti}, ``An open torque-controlled modular
  robot architecture for legged locomotion research,'' \emph{IEEE Robotics and
  Automation Letters}, vol.~5, no.~2, pp. 3650--3657, 2020.

\end{thebibliography}

\end{document}